%% file: Paper.tex
\documentclass[11pt,a4paper]{article}

\usepackage{amsmath,amssymb,amsfonts,amsthm}
\usepackage{algorithm, algpseudocode}
\usepackage{graphicx}
\usepackage{textcomp}
\usepackage[dvipsnames]{xcolor}
\usepackage{tikz}
\usepackage{pgfplots}
\usetikzlibrary{external, positioning,shapes, arrows}
\usepackage{mathtools}
\usepackage{varwidth}
\usepackage[utf8]{inputenc}
\usepackage[T1]{fontenc}
\usepackage{enumitem}
\usepackage{siunitx}
\usepackage{balance}
\usepackage{multirow}
\usepackage{booktabs}
\usepackage{caption}
\usepackage{subcaption}
\usepackage[font=small]{caption}
\usepackage{setspace}

\usepgfplotslibrary{fillbetween}
\usetikzlibrary{patterns.meta}

\usepackage[backend=biber, style=ieee, sorting=none, dashed=false]{biblatex}

\input{tex/tex_variables}

\usepackage{fancyhdr}
\fancypagestyle{FirstPage}{
	\cfoot{\footnotesize\textbf{Original publication DOI:} 10.1109/LRA.2022.3194688\\ \textbf{\copyright2022 IEEE.} Personal use of this material is permitted. Permission from IEEE must be obtained for all other uses, in any current or future media, including reprinting/republishing this material for advertising or promotional purposes, creating new collective works, for resale or redistribution to servers or lists, or reuse of any copyrighted component of this work in other works.} 
}
\begin{document}
\newbox\keywbox
\setbox\keywbox=\hbox{\bfseries Keywords:}%

\newcommand\keywords{%
	\noindent\rule{\wd\keywbox}{0.25pt}\\\textbf{Keywords:}\ }
\title{Kinematic Orienteering Problem with Time-Optimal Trajectories for Multirotor UAVs}
\author{$\text{Fabian Meyer}^{1\text{st}}$, fabian.meyer@fzi.de \and $\text{Katharina Glock}^{2\text{nd}}$, kglock@fzi.de}
\date{}

\maketitle

\begin{abstract}
	In many unmanned aerial vehicle (UAV) applications for surveillance and data collection, it is not possible to reach all requested locations due to the given maximum flight time. Hence, the requested locations must be prioritized and the problem of selecting the most important locations is modeled as an Orienteering Problem (OP). To fully exploit the kinematic properties of the UAV in such scenarios, we combine the OP with the generation of time-optimal trajectories with bounds on velocity and acceleration. We define the resulting problem as the Kinematic Orienteering Problem (KOP) and propose an exact mixed-integer formulation together with a Large Neighborhood Search (LNS) as a heuristic solution method. We demonstrate the effectiveness of our approach based on Orienteering instances from the literature and benchmark against optimal solutions of the Dubins Orienteering Problem (DOP) as the state-of-the-art. Additionally, we show by simulation \color{black} that the resulting solutions can be tracked precisely by a modern MPC-based flight controller. Since we demonstrate that the state-of-the-art in generating time-optimal trajectories in multiple dimensions is not generally correct, we further present an improved analytical method for time-optimal trajectory generation. 
	
	\keywords Orienteering Problem, multirotor UAV, route planning, trajectory generation, time-optimality

\end{abstract}

\input{tex/introduction}

\input{tex/related_work}
\input{tex/trajectory_generation_new}
\input{tex/problem_statement}

\input{tex/heuristic_approach}

\input{tex/results}
\input{tex/conclusion}

\printbibliography
\end{document}

%% file: tex/tex_variables.tex
\newcommand{\numWP}{L}
\newcommand{\setWP}{\mathcal{L}}
\newcommand{\numHead}{H}
\newcommand{\setHead}{\mathcal{H}}
\newcommand{\numVelo}{V}
\newcommand{\setVelo}{\mathcal{V}}
\newcommand{\ratio}{\mathcal{R}}

%% file: tex/introduction.tex
\section{Introduction}
\thispagestyle{FirstPage}
In the last decade, UAV technology has been steadily gaining momentum. With technological advancements, UAVs are proving to be extremely useful in a variety of applications. Within this work, we focus on flight planning for surveillance and data collection, which is one of the most important use cases for UAVs \cite{Otto.2018}. In practical applications, UAVs have a limited flight time due to their battery capacity, which is why a selection of the requested locations for data collection must often be made. To do this, locations can be prioritized and the problem can be modeled as an Orienteering Problem (OP), which aims to find the trajectory that maximizes the collected priorities given the maximum flight time constraint (see \cite{Penicka.2017, Faigl.2019, Fountoulakis.2020}). For related problems, current approaches consider the UAV's physics to generate efficient flight motions and use them as flight time estimation between locations. The most prominent examples are so-called Dubins paths and Bézier curves. Dubins paths are based on the principle of minimum turning radii, which are always feasible with regard to the underlying physics (see \cite{Otto.2018}). However, their disadvantage is that they are based on the assumption of a constant velocity. Especially for the mission planning of the widely used multirotor UAVs, the degree of freedom offered by longitudinal acceleration, which enables flying sharp turns, is thus omitted. Bézier curves, in turn, allow longitudinal acceleration. However, they do not intrinsically consider physical restrictions of the UAV such as maximum acceleration, which is why a feasible trajectory along the curve must be determined via post-processing. Furthermore, they do not provide any guarantees regarding time-efficiency.
\begin{figure}
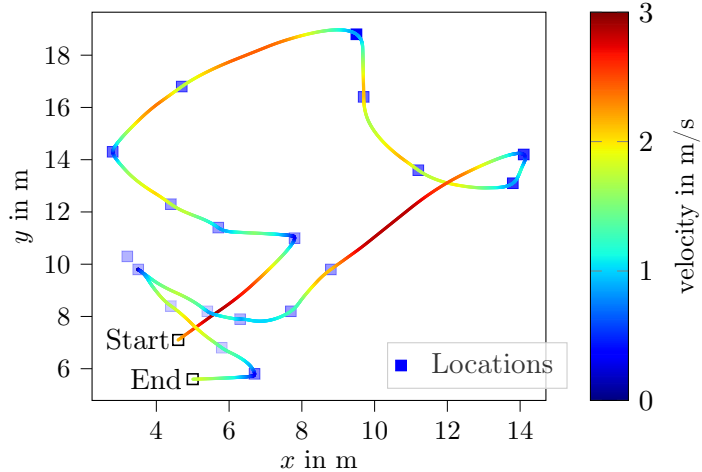

	\centering
	\include{tikz/KOPSolution}
	\vspace{-8mm}
	\caption{Example solution of the KOP for the second OP instance from Tsiligirides with bounds on the allowed acceleration $a(t) \in \left[-1.5, 1.5 \right]\left(\frac{\text{m}}{\text{s}^2}\right) $ as well as bounds on the allowed velocity $v(t) \in \left[ -3, 3\right] \left(\frac{\text{m}}{\text{s}}\right)$ and a maximum flight time $C_{max}=35$\,s.The changing color of the trajectory represents the corrensponding velocity. The priority at each location is indicated by its opacity with a high opacity referring to a high priority.}
	\vspace{-3mm}
	\label{fig:kop}
\end{figure}

For this reason, we propose a solution approach which is able to consider acceleration in arbitrary direction and guarantee physical feasibility as well as time-efficiency. Our approach combines the generation of time-optimal trajectories with bounds on velocity and acceleration, further denoted as kinematic trajectories, with the well-known Orienteering Problem and is therefore called Kinematic Orienteering Problem (KOP). For better understanding, Fig. \ref{fig:kop} shows an example solution to the problem.
With this work, we extend our previous research \cite{Meyer.2021} and make the following contributions:
\begin{itemize}
	\item We introduce the KOP as the problem of finding the kinematic trajectory that maximizes the collected priorities without exceeding the maximum allowed flight time.
	\item Further, we propose a mathematical problem formulation for the KOP and propose an easy-to-implement heuristic solution approach that yields high-quality solutions in short time.
	\item Additionally, we present an improved analytical approach to generate time-optimal kinematic trajectories in multiple dimensions, since we found that the state-of-the-art procedure is not generally correct.
	\item We benchmark our proposed solution approach against exact solution of the DOP and show by simulation that the trajectories found by our approach can precisely be tracked by a modern MPC-based flight controller.
\end{itemize}

The outline of this paper is as follows: We give an overview of related literature in Section \ref{Sec:Related_Work}. Next, we present our approach to generating time-optimal kinematic trajectories in Section \ref{Sec:TrajectoryGeneration}, followed by the mathematical definition of the KOP in Section \ref{Sec:Routing}. Section \ref{Sec:Heuristic} presents our heuristic solution approach and Section \ref{Sec:Results} shows the results of our approach. Finally, Section \ref{Sec:Conclusion} concludes and gives an outlook.

%% file: tex/related_work.tex
\section{Related Work} \label{Sec:Related_Work}
Since this work is, to our knowledge, the first to combine the classical Orienteering Problem, which is a subclass of route planning, with the generation of time-optimal kinematic trajectories for UAVs, this section is divided into two parts. In the first part, we give an overview of current approaches for UAV route planning. In the second part, we present the latest approaches for UAV trajectory generation.

\subsection{Route Planning Problems for Multirotor UAVs}
Route planning problems for multirotor UAVs are enjoying growing interest in the scientific literature. Although algorithms for solving UAV route planning problems use time-of-flight estimation based on Euclidean distances until now (see \cite{Fountoulakis.2020}) the trend has moved towards more precise and practically feasible metrics \cite{Henchey.2020}.

To do this, established methods include simple physical properties of UAV movements into route planning. One such characteristic is the minimum turning radius, which results from a UAV moving at constant velocity and applying maximum lateral acceleration. It is a restriction that comes into effect especially for fixed-wing UAVs but is also considered for multirotor UAVs. The resulting flight paths are known as Dubins paths and are proven to be optimal regarding the assumed kinematic restrictions (see \cite{Dubins.1957}). This principle is used in many UAV route planning problems. Examples include \cite{Penicka.2017} and \cite{Sundar.2022}. The advantage of using Dubins paths and thus assuming constant velocity is that acceleration and deceleration are avoided, making the trajectory more energy-efficient overall. Its disadvantage is that large detours may have to be accepted since the maximum acceleration has to fight against the prevailing mass inertia. At some point, these detours cancel out the efficiency advantage of a constant velocity and that is why an optimal tradeoff has to be determined \cite{Meyer.2021}. 

Another possibility to estimate flight times for UAV route planning problems is by using so-called Bézier curves \cite{Faigl.2018, Faigl.2019}. Here, two locations are connected by a smooth curve based on Bernstein polynomials. By setting intermediate control points in the right way, it is possible to guide the path safely around obstacles. However, the polynomial representation of a safe path through Bézier curves is purely spatial. To ensure physical feasibility, the Bézier curve must first be transferred to the time domain. Only by assigning each spatial point of the curve to a particular point in time is it possible to consider physical constraints such as maximum velocity and acceleration (see \cite{Gao.2018}). For routing problems where thousands of trajectories are calculated this two-step procedure is computationally expensive (see \cite{Faigl.2018}). Further, it is argued in \cite{Gao.2018} that if the initial and final velocity are not both zero, there may not exist a feasible solution. Moreover, another disadvantage is, especially for environments without obstacles, that the resulting trajectory is bound to the precomputed path and therefore likely to be time-suboptimal.

\subsection{Trajectory Generation for Multirotor UAVs}
Apart from Dubins paths and Bézier curves there are many trajectory generation approaches that plan UAV motions directly within the time domain and hence might suit better for flight time estimation. However, some of these approaches do not consider bounds on maximum velocity and acceleration either. An example is the well-known approach to generate minimum-snap trajectories, see \cite{Mellinger.2011, Richter.2016}.

An approach that considers bounds on velocity and acceleration is model-based predictive control (MPC). It is based on a discrete-time motion model of the UAV consisting of system parameters, system state, and control variables. The objective is to find an energy-efficient sequence of control inputs that minimizes the deviation of the current state to a reference, for example, described by the desired end state, while respecting limits of control input and feasible states \cite{Mueller.2013, Kamel.2017}. However, MPC does not provide time-optimality, and since MPC solves a mathematical program, usually quadratic, numerically, it is a computationally intensive method. 

The last form of trajectory optimization addressed here deals with the determination of time-optimal trajectories \cite{Beul.2016, Beul.2017}. According to Pontryagin's minimum principle (see \cite{Pontryagin.1987}), time-optimality is achieved by having the system always operate at its physical limits. For this purpose, the overall motion of the UAV is divided into several time segments and for each, the control variable may be the maximum or the minimum control value or zero. This property is used to calculate the trajectories analytically.
Further, \cite{Beul.2016} argues that time- and energy-optimality are equal in terms of the total thrust integral. Faster trajectories might consume more energy for acceleration and deceleration, but since their flight time is shorter, their energy consumption is less compared to slower trajectories.

In our previous work \cite{Meyer.2021}, we adapt this approach by using acceleration as control input. This enables us to generate trajectories considering fundamental physical properties and to obtain high-quality travel time estimates in a very short time. However, we found that the state-of-the-art procedure that we based our trajectory generation on is not generally correct. In Section \ref{Sec:TrajectoryGeneration}, we give an example for our observation and present a refined method that solves that issue.

%% file: tex/trajectory_generation_new.tex
\section{Trajectory Generation} \label{Sec:TrajectoryGeneration}
The state-of-the-art to determine time-optimal trajectories from an arbitrary initial state to an arbitrary final state with constraints on system state and control input is based on the decoupling of axes (see \cite{Beul.2017, Mueller2.2013, Hehn.2015}). This means that in a first step all spatial coordinate axes $i \in \left\{1,..., n\right\}$ of the trajectory in the $n$-dimensional space are separated. Pontryagin's minimum principle is applied for each axis yielding a bang-zero-bang control input pattern. For the $i$-th axis with bounded acceleration as control input and a maximum allowed velocity, the resulting bang-zero-bang acceleration pattern $(a_1, a_2, a_3)$, which defines the acceleration profile 
\begin{equation*}
	a_i(t) = \begin{cases}
		a_1, & 0 \leq t < t_{1, i} \\
		a_2, & t_{1, i} \leq t < t_{1, i}+t_{2, i} \\
		a_3, & t_{1, i}+t_{2, i} \leq t \leq t_{1, i}+t_{2, i}+t_{3, i} = t_e,\\
	\end{cases}
\end{equation*} 
is given by $(+a, 0, -a)$  with either $a = -a_{max}$ or $a = +a_{max}$  and $a_{max}$ representing the maximum allowed acceleration. The durations of each time segment of constant acceleration are described by $t_{1, i},t_{2, i}$ and $t_{3, i}$. In case the velocity limit is not reached, $t_{2, i} = 0$ holds. Consequently, the time-optimal trajectory consists of two time segments of constant maximum acceleration with opposite sign and, if the velocity limit is reached, one segment of no acceleration. The time-optimal trajectory duration $T_{opt,i}$ can be calculated analytically (e.g. see \cite{Meyer.2021}). Next, the state-of-the-art postulates that the overall duration $T_{sync}$ is defined as the largest of the time-optimal durations $T_{opt,i}$ over all axes, i.e.
\begin{equation}
	T_{sync} = \max_{i\in \left\{1, ...,n\right\}}\lbrace T_{opt, i} \rbrace. \label{eq:best_traj_time}
\end{equation}
However, this procedure sometimes results in unexpected behavior since not all axes can be synchronized with the duration $T_{sync}$. The reason is that this approach does not consider the inertia of the movement properly, and therefore, sometimes results in overshooting the desired final state. This effect has not been reported in the literature so far. Next, we give an example where this approach is invalid.

\begin{figure}[tb]
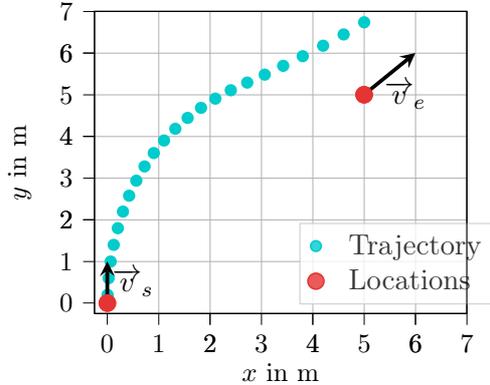

	\centering
	\include{tikz/insync}
	\vspace{-8mm}
	\caption{Example for insynchronizability of a given initial and end state.}
	\vspace{-3mm}
	\label{fig:insync}
\end{figure}

Figure \ref{fig:insync} illustrates  the trajectory generation in two dimensions $x$ and $y$ based on the state-of-the-art. For the $x$ axis the initial state is  $p_{x, s} = 0$\,m, $v_{x,s} = 0\,\frac{\text{m}}{\text{s}}$, where $p_{x, s}$ describes the initial position and $v_{x,s}$ the initial velocity. The desired end state for the $x$ axis is $p_{x, e} = 5$\,m, $v_{x,e}=2\,\frac{\text{m}}{\text{s}}$. The time-optimal duration  for a maximum allowed acceleration $a \in \left[-0.5, 0.5\right]\,(\frac{\text{m}}{\text{s}^2})$ and velocity  $v \in \left[-2, 2\right]\,(\frac{\text{m}}{\text{s}})$ can be calculated as described in \cite{Meyer.2021} and is $T_{opt, x} = 4.5$\,s. For the $y$ axis the initial and end state are $p_{y, s} = 0$\,m, $v_{y,s} = 2\,\frac{\text{m}}{\text{s}}$, $p_{y, e} = 5$\,m, $v_{y,e}=2\,\frac{\text{m}}{\text{s}}$. Here, the calculation of the time-optimal duration yields $T_{opt, y} = 2.5$\,s. According to Equation \eqref{eq:best_traj_time}, both axes must be synchronized at $T_{sync} = 4.5$\,s. By utilizing MPC for trajectory synchronization with a fixed duration, it can be seen that the resulting two dimensional trajectory (blue dots in Figure \ref{fig:insync}) misses the required end state by far. This behavior is due to the inertia of the system. The high initial velocity along the $y$ axis in combination with an insufficient acceleration power leads to overshooting the end position $p_{y, e}$. As a result, although the $y$ axis has the potentially faster execution, it cannot be synchronized with the slower time-optimal duration of the $x$ axis.

\begin{figure}
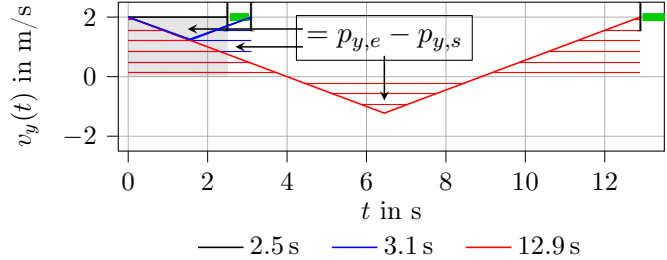

	\centering
	\include{tikz/insync2}
	\vspace{-8mm}
	\caption{Range of valid trajectory durations shown in green. The velocity profiles that lead to the limits of the corresponding range are given by colored lines. The associated areas underneath these profiles are highlighted in the respective colors.}
	\vspace{-3mm}
	\label{fig:insync2}
\end{figure}
For the above scenario, Figure \ref{fig:insync2} shows feasible velocity profiles for different synchronization times $T_{sync}$ for the $y$ axis with respect to the desired end state. The property of a feasible trajectory is that the desired start and end velocity must be achieved, and the integral underneath the velocity profile must equal $p_{y,e} - p_{y,s}$. The velocity profile $v_y(t)$ of the time-optimal trajectory is shown by a solid black line. Since its initial velocity already equals the maximum allowed value, it remains constant and $T_{y} = T_{opt,y} = 2.5$\,s results. If it is required to increase the trajectory duration, i.e.  $T_{y} > 2.5$\,s, the UAV first decelerates the motion and accelerates afterwards to meet the required final velocity $v_{y, e} = 2\,\frac{\text{m}}{\text{s}}$. However, this procedure works only as long as the integral underneath the velocity profile equals $p_{y,e} - p_{y,s}$. The slowest velocity profile that does not violate this condition corresponds to $T_{y} \approx \text{3.1\,s}$(see blue line). Further increasing the required trajectory duration leads to overshooting until $T_{y} \geq \text{12.9\,s}$(see red line). From this duration on it is possible to compensate for overshooting by flying a turn. In total, the range of feasible trajectory durations with respect to the requirements on the $y$ axis is given in green. As can be seen, a synchronization with $T_{sync} = T_{opt, x} = 4.5$\,s is not possible.

In the following, we show our general approach to design time-optimal trajectories for multiple axes. First, we present acceleration patterns needed for our approach and discuss how we can check whether these patterns can be synchronized with a particular trajectory duration $T_{sync}$. Second, we describe a general procedure to obtain the time-optimal duration.

\subsection{Synchronization Feasibility}
\begin{figure}
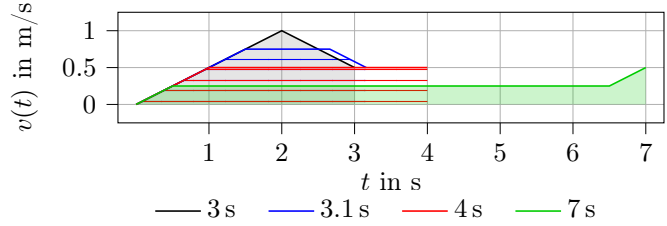

	\centering
	\include{tikz/insync3}
	\vspace{-8mm}
	\caption{Example of velocity profiles for different acceleration patterns and trajectory durations. }
	\vspace{-3mm}
	\label{fig:insync3}
\end{figure}

In this subsection, we only focus on the feasibility of synchronizing a single axis with a particular $T_{sync}$, and therefore the subscript $i$ will be discarded. To check feasibility, we first define the considered acceleration patterns. On the one hand, we consider the acceleration patterns for time-optimality in a single axis given by $(+a, 0, -a)$  with $a \in \lbrace -a_{max}, a_{max}\rbrace$. These patterns are from now on called classical patterns. However, we further consider the patterns defined by $(+a, 0, +a)$  with $a \in \lbrace -a_{max}, a_{max}\rbrace$, which we denote as synchronization patterns. The reason for the latter is that classical patterns aim at finding the time-optimal behavior, however, they are not sufficient in some cases of synchronization with large synchronization times. We give the following example for illustration (see Figure \ref{fig:insync3}): It is assumed that an axis with $p_s = 0\,\text{m}, p_e = \text{$1.75$\,m}$and $v_s = 0\,\frac{\text{m}}{\text{s}}, v_e =0.5\,\frac{\text{m}}{\text{s}}$ has to be synchronized with a large enough duration $T_{sync}$. Here, $p_s$, $p_e$, $v_s$ and $v_e$ describe initial and end position as well as velocity. 

The pattern yielding time-optimality is given by $(+a_{max}, 0, -a_{max})$, with the black line giving the overall time-optimal velocity profile. However, this pattern is only applicable as long as $T_{sync} \leq 4$\,s. When $T_{sync}$ is increased, the segment of deceleration at the end of the velocity profile shortens in time (blue line) until it becomes zero for synchronization time $T_{sync} = 4$ (red line). If it is required to synchronize the axis with $T_{sync} > 4$\,s using the classical pattern $(+a_{max}, 0, -a_{max})$, no solution can be found anymore since it is not possible to meet the desired final velocity and area underneath the velocity profile at the same time. In this case, the synchronization patterns $(+a, 0, +a)$, $a \in \lbrace -a_{max}, a_{max}\rbrace$ with two phases of acceleration pointing towards the same direction come into play (green line). With such a pattern it is possible to synchronize the axis with $T_{sync} > 4$\,s. We define these patterns as synchronization patterns because they are only needed for axis synchronization.  

To guarantee synchronization feasibility of one axis with the trajectory time $T_{sync}$, one has to find a pattern from the set of classical and synchronization patterns where $t_1, t_2, t_3 \geq 0$\,s and $v_{min} \leq v(t) \leq v_{max}, \forall t\in \left[0, T_{sync}\right]$. We define these inequations as synchronization conditions. Note that, if the initial and end velocities $v_s$ and $v_e$ are within the velocity bounds, it is sufficient to show that the constant velocity $v_c=v(t_1)$ in the segment of no acceleration is within the velocity bound to guarantee that $v_{min} \leq v(t) \leq v_{max}$.  The following subsections present how the values $t_1, t_2, t_3, v_c$ are determined for classical and synchronization patterns.

\subsubsection{Classical Patterns} \label{sec:optimality_pattern}
The classical patterns are defined by the acceleration profile
\begin{equation}
	a(t) = \begin{cases}
		+a, & 0 \leq t < t_1 \\
		0, & t_1 \leq t < t_1+t_2\\
		-a, & t_1+t_2 \leq t \leq t_1+t_2+t_3.
	\end{cases}  \label{eq:acc_profile_opt}
\end{equation}
with $a \in \lbrace a_{max}, -a_{max}\rbrace$. Based on this acceleration profile, the velocity profile results in 
\begin{equation}
	v(t) = \begin{cases}
		v_s + a t, & 0 \leq t < t_1 \\
		v_s + a t_1, & t_1 \leq t < t_1+t_2 \\
		v_s + 2a t_1 - a\text{$(t - t_2)$} & t_1+t_2 \leq t \leq t_1+t_2+t_3
	\end{cases}\label{eq:vel_profile_opt}
\end{equation}
In order to meet the required velocity $v_e$ at time $t_1 + t_2 + t_3$ the following equation has to hold:
\begin{align}\label{eq:vel_constraint_opt}
	v_e - v_s &= \int_{0}^{t_1 + t_2 + t_3}a(t)\text{d}t \nonumber\\
	&= at_1 - at_3
\end{align}
To meet the required position $p_e$ at time $t_1 + t_2 + t_3$ the equation 
\begin{align}\label{eq:pos_constraint_opt}
	p_e - p_s &= \int_{0}^{t_1 + t_2 + t_3}v(t)\text{d}t \nonumber\\
	&=v_st_1+\frac{1}{2}at_1^2+(v_s+at_1)t_2+(v_s+at_1)t_3-\frac{1}{2}at_3^2 
\end{align}
has to hold as well. 
Additionally, the equation
\begin{equation}\label{eq:t_constraint_opt}
	t_1 + t_2 + t_3 = T_{sync}
\end{equation}
has to hold to guarantee time synchronization. Further, since the velocity $v_c$ for $t \in \left[t_1, t_2\right]$ is constant, the equation 
\begin{equation}\label{eq:vel_limit}
	v_c = at_1 +v_s 
\end{equation}
applies as well.

Equations \eqref{eq:vel_constraint_opt}, \eqref{eq:pos_constraint_opt}, \eqref{eq:t_constraint_opt} and \eqref{eq:vel_limit} form a system of equations with variables $t_1, t_2, t_3, v_c$ whose solution is given by
\begin{align}
	t_1 &= \frac{aT_{sync}+v_e-v_s \pm\sqrt{A}}{2a} \label{eq:t_1_opt}\\ 
	t_2 &= \mp\frac{\sqrt{A}}{a}\label{eq:t_2_opt}\\
	t_3 &= \frac{aT_{sync}-v_e+v_s \pm\sqrt{A}}{2a}\label{eq:t_3_opt}\\
	v_c &=\frac{aT_{sync} + v_s + v_e + \sqrt{A}}{2} \label{eq:v_c}
\end{align}
with 
\begin{equation}
	A= a^2T_{sync}^2+2(v_e+v_s)aT_{sync}-4a(x_e-x_s)-(v_e-v_s)^2.
\end{equation}
Equations \eqref{eq:t_1_opt}, \eqref{eq:t_2_opt}, \eqref{eq:t_3_opt} and \eqref{eq:v_c} only depend on the trajectory duration $T_{sync}$. Hence, it is sufficient to insert $T_{sync}$ into these equations and verify that $t_1(T_{sync}), t_2(T_{sync}), t_3(T_{sync}) \geq 0$ and $v_{min}\leq v_c(T_{sync}) \leq v_{max}$ to show synchronization feasibility for the respective classical pattern.   

\subsubsection{Synchronization Patterns}
Synchronization feasibility for the synchronization patterns is checked analogously. The only difference is in the applied acceleration pattern 
\begin{equation}
	a(t) = \begin{cases}
		+a, & 0 \leq t < t_1 \\
		0, & t_1 \leq t < t_1+t_2\\
		+a, & t_1+t_2 \leq t \leq t_1+t_2+t_3
	\end{cases}  \label{eq:acc_profile_sync}
\end{equation}
with $a \in \lbrace a_{max}, -a_{max}\rbrace$, which results in the following velocity profile
\begin{equation}
	v(t) = \begin{cases}
		v_s + a t, & 0 \leq t < t_1 \\
		v_s + a t_1, & t_1 \leq t < t_1+t_2 \\
		v_s + a\text{$(t - t_2)$} & t_1+t_2 \leq t \leq t_1+t_2+t_3
	\end{cases}\label{eq:vel_profile_sync}
\end{equation}
Analogously, this leads to a system of four equations and four variables whose solution is given by 
\begin{align}
	t_1 &= \frac{(-2v_sT_{sync} + 2(p_e-p_s))a - (v_e-v_s)^2}{2a(T_{sync}a-v_e+v_s)}\label{eq:t_1_sync}\\ 
	t_2 &= \frac{aT_{sync}-v_e+v_s}{a}\label{eq:t_2_sync}\\ 
	t_3 &= \frac{(-2v_eT_{sync} - 2(p_e-p_s))a - (v_e-v_s)^2}{2a(T_{sync}a-v_e+v_s)}\label{eq:t_3_sync}\\
	v_c &= \frac{2a(p_e - p_s) - v_e^2 + v_s^2}{2(aT_{sync} - v_e + v_s)} \label{eq:vc_sync}
\end{align}
Again, equations \eqref{eq:t_1_sync}, \eqref{eq:t_2_sync}, \eqref{eq:t_3_sync} and \eqref{eq:vc_sync} only depend on $T_{sync}$ and the synchronization conditions can easily be checked via insertion.  

\subsection{Optimal Synchronization Time}
In this subsection, we show how to find the time-optimal duration for multiple axes. The set of valid synchronization times $\Omega$ is continuous and constrained by the synchronization conditions. To find the time-optimal duration for multiple axes, one has to investigate the boundary of $\Omega$. This is where $t_1, t_2, t_3 \geq 0$ as well as $v_c \geq v_{min}$ and $v_c \leq v_{max}$ holds and at least one of these inequalities is fulfilled with equality. Hence, each $T_{sync}$ that fulfills equality for any of the synchronization conditions is a potential candidate for the time-optimal duration for multiple axes. To determine the best synchronization time for multiple axes, all candidates $T_{sync}\geq \max_{i\in \left\{1, ...,n\right\}}\lbrace T_{opt, i} \rbrace$, with $T_{opt, i}$ for the $i$-th axis calculated as described in \cite{Meyer.2021}, are inserted into the synchronization conditions of each axis and pattern and checked for feasibility. The lowest trajectory synchronization time that yields feasibility for at least one pattern in each axis is defined as the optimal synchronization time $T^\ast$. With the associated values $t_1, t_2, t_3$ for the corresponding patterns it is possible to reconstruct the trajectory.

%% file: tikz/insync.tex
\begin{tikzpicture}
\definecolor{traj}{RGB}{0,204,204}
\definecolor{red}{rgb}{0.9,0.2,0.2}%

\begin{axis}[height = 2.2in,
legend cell align={left},
legend style={
  fill opacity=0.8,
  draw opacity=1,
  text opacity=1,
  at={(0.55,0.3)},
  anchor=north west,
  draw=white!80!black
},
tick align=outside,
tick pos=left,
x grid style={white!69.0196078431373!black},
xlabel={\small $x$ in m},
xmin=-0.25, xmax=7,
xtick style={color=black},
xticklabel style = {font= \small},
xtick={1, 2, 3, 4, 5, 6, 7},
y grid style={white!69.0196078431373!black},
ylabel={\small $y$ in m},
y label style = {at={(0.05,0.5)}},
ymin=-0.25, ymax=7,
ytick style={color=black},
yticklabel style = {font= \small},
ytick={1, 2, 3, 4, 5, 6, 7},
grid=both,
grid style={line width=.1pt, draw=gray!10},
major grid style={line width=.2pt,draw=gray!50}
]
\addplot [semithick, traj, mark=*, mark size=2, mark options={solid}, only marks]
table {%
0.00250000000000009 0.199886596462648
0.0224999999999975 0.600048620833216
0.0624999999999903 1.00010853066118
0.122499999999979 1.3999774310263
0.202499999999963 1.80018996606698
0.302499999999942 2.19850639850034
0.422499999999918 2.57964168574598
0.562499999999888 2.9407983062079
0.722499999999855 3.2819737161916
0.902499999999817 3.60316539101956
1.10250000000001 3.90437084653992
1.32250000000049 4.18558766043709
1.56250000000106 4.4468134933433
1.82250000000172 4.68804610977221
2.10250000000246 4.90928339891561
2.40250000000328 5.11052339535954
2.72250000000419 5.29423041083401
3.06250000000518 5.4853353632961
3.42250000000626 5.6964394709783
3.80250000000742 5.92754253763613
4.19999499750777 6.17864453265532
4.59998499750724 6.44974557490197
4.99997499750671 6.74084591667184
};
\addlegendentry{Trajectory}

\addplot [semithick, red, mark=*, mark size=3, mark options={solid}, only marks]
table {%
	0 0
	5 5
};
\addlegendentry{Locations}

\end{axis}
\begin{axis}[height = 2.2in,
	legend cell align={left},
	legend style={
		fill opacity=0.8,
		draw opacity=1,
		text opacity=1,
		at={(0.03,0.97)},
		anchor=north west,
		draw=white!80!black
	},
	tick align=outside,
	tick pos=left,
	x grid style={white!69.0196078431373!black},
	xmin=-0.25, xmax=7,
	xtick style={color=black},
	xticklabel style = {font= \small},
	y grid style={white!69.0196078431373!black},
	ymin=-0.25, ymax=7,
	ytick style={color=black},
	yticklabel style = {font= \small},	]
\addplot [semithick, red, mark=*, mark size=3, mark options={solid}, only marks]
table {%
	0 0
	5 5
};
	
\draw [->,-stealth, line width=0.5mm] (axis cs:0,0) -- (axis cs:0, 1);
\draw [->,-stealth, line width=0.5mm] (axis cs:5,5) -- (axis cs: 6,6);
\node[] (A) at(axis cs:0.5, 0.5) {$\overrightarrow{v}_s$};
\node[] (A) at (axis cs:5.8,5.0 ) {$\overrightarrow{v}_e$}; 
\end{axis}
\end{tikzpicture}

%% file: tikz/insync2.tex
\usepgfplotslibrary{fillbetween} 
\usetikzlibrary{patterns.meta}
\begin{tikzpicture}
	\definecolor{black}{RGB}{0,0,0}
	\definecolor{traj1}{RGB}{0,0,204}
	\definecolor{traj2}{RGB}{0,0,204}
	\definecolor{traj3}{rgb}{0.9,0.2,0.2}%
	\definecolor{traj4}{RGB}{0,204,0}

	\begin{axis}[height = 1.5in, 
		width = 3.45in,
		height=1.4in,
		legend cell align={left},
		legend style={
			at={(0.5,-0.5)},
			anchor=north,
			draw=none
		},
		legend columns=-1,
		tick align=outside,
		tick pos=left,
		x grid style={white!69.0196078431373!black},
		xlabel={\small $t$ in s},
		xmin=-0.25, xmax=13.5,
		xtick style={color=black},
		xticklabel style = {font= \small},
		y grid style={white!69.0196078431373!black},
		ylabel={\small $v_y(t)$ in m/s},
		ymin=-2.5, ymax=2.5,
		ytick style={color=black},
		yticklabel style = {font= \small},
		grid=both,
		grid style={line width=.1pt, draw=gray!10},
		major grid style={line width=.2pt,draw=gray!50}
		]
		\addplot [name path = p1, semithick, black]
		table {%
			0 2
			2.5 2
			
		};
		\addlegendentry{\small $2.5$\,s$\quad$}

		\addplot [name path = p2, semithick, blue ]
		table {%
			0 2
			1.55 1.225
			3.1 2
			
		};
		\addlegendentry{\small $3.1$\,s$\quad$}
		
		\addplot [name path = p3, semithick, red]
		table {%
			0 2
			6.449 -1.225
			12.89 2
			
		};
		\addlegendentry{\small $12.9$\,s}
		
		\addplot[name path=X1,draw=none]
		table {%
			0 0
			2.5 0	
		};
		
		\addplot[name path=X2,draw=none]
		table {%
			0 0
			3.1 0	
		};
		
		\addplot[name path=X3,draw=none]
		table {%
			0 0
		12.89 0	
		};
		\addplot [gray, opacity=0.2] fill between[ 
		of=p1 and X1
		];

		\addplot [pattern={Lines[angle=-45,distance=4pt]}, pattern color=blue] fill between[ 
		of=p2 and X2
		];
		\addplot[pattern={Lines[angle=45,distance=4pt]}, pattern color=red] fill between[of=p3 and X3];
		
		\node (one) at (axis cs:2.5,2) {]};
		\node (two) at (axis cs:3.1,2) {[};
		\node (three) at (axis cs:12.9,2) {]};
		\node[draw] (four) at (axis cs:6.45,1.3) {$= p_{y,e} - p_{y,s}$};
		
		\draw [->,stealth-, line width=0.2mm](axis cs:6.45,-0.8) -- (axis cs:6.45,0.7);
		\draw [->,stealth-, line width=0.2mm](axis cs:2.7,1) -- (axis cs:4.4,1);
		\draw [->,stealth-, line width=0.2mm](axis cs:1.5,1.6) -- (axis cs:4.4,1.6);

		\addplot [line width=3pt, traj4]
		table {%
			2.56 2
			3.04 2
		};
		
		\addplot [line width=3pt, traj4]
		table {%
			12.95 2
			17 2
		};

		\addplot [name path = p2, semithick, blue]
		table {%
			0 2
			1.55 1.255
			3.1 2
			
		};
	\end{axis}
	
\end{tikzpicture}

%% file: tikz/insync3.tex
\begin{tikzpicture}
	\definecolor{black}{RGB}{0,0,0}
	\definecolor{traj1}{RGB}{0,204,204}
	\definecolor{traj2}{RGB}{0,0,204}
	\definecolor{traj4}{RGB}{0,204,0}
	\definecolor{traj3}{rgb}{0.9,0.2,0.2}%

	\begin{axis}[height = 1.3in, 
		width = 3.45in,
		height=1.2in,
		legend cell align={left},
		legend style={
			at={(0.5,-0.6)},
			anchor=north,
			draw=none
		},
		legend columns=-1,
		tick align=outside,
		tick pos=left,
		x grid style={white!69.0196078431373!black},
		xlabel={\small $t$ in s},
		xmin=-0.25, xmax=7.25,
		xtick style={color=black},
		xticklabel style = {font= \small},
		xtick={1, 2, 3, 4, 5, 6, 7},
		y grid style={white!69.0196078431373!black},
		ylabel={\small $v(t)$ in m/s},
		ymin=-0.25, ymax=1.25,
		ytick style={color=black},
		yticklabel style = {font= \small},
		grid=both,
		grid style={line width=.1pt, draw=gray!10},
		major grid style={line width=.2pt,draw=gray!50}
		]
		
%
%
%
%

		\addplot [name path=P1, semithick, black]
		table {%
			0 0
			2 1
			3 0.5
		};
		\addlegendentry{\small $3$\,s$\quad$}

		\addplot [name path=P2, semithick, blue]
		table {%
			0 0
			1.5 0.75
			2.66 0.75
			3.16 0.5
		};
		\addlegendentry{\small $3.1$\,s$\quad$}
		
		\addplot [name path=P3, semithick, red]
		table {%
			0 0
			1 0.5
			4 0.5
		};
		\addlegendentry{\small $4$\,s$\quad$}
		
		\addplot [name path=P4, semithick, traj4]
		table {%
			0 0
			0.5 0.25
			6.5 0.25
			7 0.5
		};
		\addlegendentry{\small $7$\,s$\quad$}

		\addplot[name path=X1,draw=none]
		table {%
			0 0
			3 0	
		};
	
		\addplot[name path=X2,draw=none]
		table {%
			0 0
			3.16 0	
		};
			
		\addplot[name path=X3,draw=none]
		table {%
			0 0
			4 0	
		};
	
		\addplot[name path=X4,draw=none]
		table {%
			0 0
			7 0	
		};

		\addplot [gray, opacity=0.2] fill between[of=P1 and X1];
		\addplot [pattern={Lines[angle=-45,distance=4pt]}, pattern color=blue] fill between[ 
		of=P2 and X2
		];
		\addplot[pattern={Lines[angle=45,distance=4pt]}, pattern color=red] fill between[of=P3 and X3];
		\addplot [traj4, opacity=0.2] fill between[ 
		of=P4 and X4
		];
		
	\end{axis}
	
\end{tikzpicture}

%% file: tex/problem_statement.tex
\section{Kinematic Orienteering Problem}\label{Sec:Routing}

The objective of the KOP is to find the kinematic trajectory through a set of locations that maximizes the sum of priorities of the selected locations  while considering the maximum flight time. The proposed mathematical optimization model for the KOP assumes that a UAV can move through a location with a discretized heading angle and additionally at a discretized velocity. As a generalization of the OP, the KOP is also NP-hard (see \cite{Golden1987}).
\subsection{Assumptions and Notations}
We solve the KOP for multirotor UAVs in an obstacle-free two-dimensional plane where each element in a set of locations $\setWP = \lbrace l_i: i = 1,...,  \numWP\rbrace$ can be traversed with discretized heading angle and with a discretized velocity and is assigned a priority $r_i \in \mathbb{R}_0^+$. Start and end locations are represented by $l_1$ and $l_{\numWP}$. The two-dimensional case is addressed, since it is the basis of an extension into the three-dimensional space and to enable a direct comparison to the Dubins Orienteering Problem (DOP) as state-of-the-art on benchmark instances.

With $\numHead$ and $\numVelo$ representing the number of discretization levels for heading angles and velocities, we define the set of discretized heading angles as $\setHead = \left\{h_i: h_i = 2\pi i/\numHead,  i = 1,...,\numHead\right\}$. The set of discretized velocities depends on the maximum allowed velocity and consists of elements $\setVelo = \lbrace v_i: v_i = (i-1)v_{max}/(\numVelo-1), i = 1,..., \numVelo\rbrace$.

Costs that describe the flight time to get from location $l_i$ with heading angle $h_k$ and velocity $v_g$ to location $l_j$ with heading angle $h_m$ and velocity $v_\text{$w$}$ are defined by $c_{ikg}^{jm\text{$w$}}$ and are determined by our trajectory generation method presented above. The maximum flight time is denoted as $C_{max}$.

\subsection{Mathematical Formulation of the KOP}
The mathematical programming formulation to determine the priority-maximizing sequence of locations, as well as the corresponding heading angles and velocity configurations that define the UAV's trajectory, is presented in the following.

The main decision variables for our formulation $x_{ikg}^{jm\text{$w$}}$ are binary and interpreted as
\begin{equation*}
	x_{ikg}^{jm\text{$w$}} = \begin{cases}
		1, & \text{if location $l_i$ is left with heading angle $h_k$}\\
		& \text{and velocity $v_g$ towards location $l_j$, which is}\\
		& \text{entered with heading angle $h_m$ and velocity $v_\text{$w$}$},\\
		0, & \text{otherwise}
	\end{cases}
\end{equation*}
Furthermore, integer decision variables $u_i \in \left\{ 1, ..., \numWP \right\}, i=1,..., \numWP$ define the sequence of visited locations $l_i$ in the tour. The overall KOP model is given as follows.

\begin{align}
	\max & \sum_{i=2}^{\numWP-1}\sum_{j=2}^{\numWP} \sum_{k=1}^{\numHead}\sum_{m=1}^{\numHead}\sum_{g=1}^{\numVelo} \sum_{w=1}^{\numVelo} x_{ikg}^{jmw}r_j \label{OP_objective} \\
	\text{s.t.} \nonumber\\
	& \sum_{j=2}^{\numWP} \sum_{k=1}^{\numHead}\sum_{m=1}^{\numHead}\sum_{g=1}^{\numVelo}\sum_{w=1}^{\numVelo} x_{1kg}^{jmw} = 1 \label{start_depot_left_v_zero}\\
	& \sum_{i=1}^{\numWP-1} \sum_{k=1}^{\numHead}\sum_{m=1}^{\numHead}\sum_{g=1}^{\numVelo}\sum_{w=1}^{\numVelo} x_{ikg}^{\numWP mw} = 1 \label{end_depot_entered_v_zero}\\
	& \sum_{i=1}^{\numWP-1} \sum_{k=1}^{\numHead}\sum_{m=1}^{\numHead}\sum_{g=1}^{\numVelo}\sum_{w=1}^{\numVelo} x_{ikg}^{jmw} \leq 1 \quad\quad \forall j = 2, ..., \numWP\label{all_locations_only_once}\\
	& \sum_{i=1}^{\numWP-1} \sum_{k=1}^{\numHead}\sum_{g=1}^{\numVelo}x_{ikg}^{jmw} = \sum_{o=2}^{\numWP} \sum_{p=1}^{\numHead}\sum_{q=1}^{\numVelo}x_{jmw}^{opq} \nonumber\\
	& \qquad\qquad\qquad\qquad\qquad\qquad  \forall j=2,\cdots, \numWP-1 \nonumber\\
	& \qquad\qquad\qquad\qquad\qquad\qquad  \forall m=1,\cdots,\numHead \nonumber\\
	& \qquad\qquad\qquad\qquad\qquad\qquad  \forall w=1,\cdots,\numVelo\label{flow_conservation}\\
	& \sum_{i=1}^{\numWP-1}\sum_{j=2}^{\numWP} \sum_{k=1}^{\numHead}\sum_{m=1}^{\numHead}\sum_{g=1}^{\numVelo} \sum_{w=1}^{\numVelo} x_{ikg}^{jmw}c_{ikg}^{jmw} \leq C_{max}\label{maximum_flight_time}\\
\end{align}
\begin{align}
	& u_i-u_j + 1 \leq (\numWP-1)\left(1-\sum_{k=1}^{\numHead}\sum_{m=1}^{\numHead}\sum_{g=1}^{\numVelo}\sum_{w=1}^{\numVelo}x_{ikg}^{jmw}\right)\nonumber\\
	&\qquad\qquad\qquad\qquad\qquad\qquad \forall i,j=1,\cdots, \numWP\label{subtour_prevention_2}\\
	& u_1 = 1 \label{subtour_prevention_3}\\
	& u_i \in \lbrace2, ..., \numWP\rbrace \qquad\qquad\qquad  \forall i=2,\cdots, \numWP \label{subtour_prevention_1}\\\
	& x_{ikg}^{jmw} \in \left\lbrace 0,1 \right\rbrace \qquad\qquad\qquad  \forall i,j=1,\cdots, \numWP \nonumber\\
	& \qquad\qquad\qquad\qquad\qquad\qquad\,  \forall k,m=1,\cdots,\numHead \nonumber\\
	& \qquad\qquad\qquad\qquad\qquad\qquad \,  \forall g,w=1,\cdots,\numVelo \label{binary}
\end{align}

The objective of the presented mathematical programming formulation \eqref{OP_objective} is to maximize the collected priorities. Constraint \eqref{start_depot_left_v_zero} enforces that the start location is left whereas constraint \eqref{end_depot_entered_v_zero} enforces that the end location is entered. Constraint set \eqref{all_locations_only_once} ensures that each location is entered at most once. Flow conservation is fulfilled by constraint set \eqref{flow_conservation}. These constraints ensure that location $l_j$ is left in the same direction as it is entered as well as with the same velocity. Constraints \eqref{maximum_flight_time} ensure that the maximum allowed flight time is not exceeded. To prevent subtours, we make use of the subtour elimination constraints \eqref{subtour_prevention_2}, \eqref{subtour_prevention_3} and \eqref{subtour_prevention_1} which are formulated according to the Miller-Tucker-Zemlin (MTZ) formulation \cite{Miller.1960}. Constraints \eqref{binary} enforce the decision variable $x_{ikg}^{jmw}$ to be binary.

%% file: tex/heuristic_approach.tex
\section{Heuristic Solution Approach} \label{Sec:Heuristic}
For larger problem instances, the KOP is too complex to be solved exactly. Therefore, we propose a heuristic solution approach based on a Large Neighborhood Search (LNS) framework, which is a widely used and easy-to-implement approach for route planning problems \cite{Pisinger.2018}. Starting from an initial solution generated by a construction heuristic, we iteratively destroy 50\% of the solution and then apply the construction heuristic again. After 100 iterations, we use the best-found solution as a new initial solution for the same procedure for another 100 iterations but destroy only 20\% of the solution. The objective is to search for good solutions globally in the first step, and in the second step, we are locally optimizing the best-found solution so far. In the following, we present the applied construction and the destruction heuristics.

\subsection{Construction Heuristic}\label{Subsec:Construction_Heuristic}
We propose a construction heuristic that inserts unscheduled locations $l_p$ into an existing plan by evaluating the ratio of potentially collected priority $r_p$ and additional flight time used for insertion. Since a location $l_p$ can be traversed in multiple ways according to heading angle $h_p$ and velocity $v_p$, we only consider the best insertion possibility, i.e. where the additional flight time is minimal:
\begin{equation*}
	c^\ast_{i,p,j}:=\min_{h_p \in \setHead, v_p \in \setVelo}c_{i, h_i, v_i}^{p, h_p, v_p} + c_{p, h_p, v_p}^{j, h_j, v_j} - c_{i, h_i, v_i}^{j, h_j, v_j}
\end{equation*}
In this term, $i$ and $j$ refer to the predecessor and successor locations $l_i$ and $l_j$ of location $l_p$. At this point, heading angles and velocities at the predecessor and successor are fixed. Hence, the term describes the additional costs of inserting location $l_p$ with optimal heading angle $h_p$ and velocity $v_p$ in between $l_i$ and $l_j$. The best insertion for location $l_p$ in between any predecessor $l_i$ and successor $l_j$ is associated with the highest ratio \begin{equation}\label{eq:ratio}
	\ratio_{p}^\ast = r_p \left[\min_{i, j} c^\ast_{i,p,j} \right]^{-1}. \nonumber
\end{equation}
This is done for all unscheduled locations. The insertion with the overall highest ratio is realized. The construction heuristic terminates when no further insertion is possible without violating the maximum flight time restriction. As post-processing for each insertion, we optimize the heading angle and velocity of the start and end location.
 
\subsection{Destruction Heuristics}
To destroy a part of the incumbent solution, we make use of three different destruction heuristics, which are applied randomly until the predefined percentage of destruction is reached. The first heuristic removes the location from the existing solution whose ratio between gained priority and flight time used is the lowest. As a second destruction heuristic, we remove the location whose assigned heading angle and velocity do not connect it with its predecessor and successor optimally. This is checked via full enumeration. The third heuristic combines both and removes the location whose ratio between the gained priority and the difference between current flight time and best possible flight time is the lowest.

%% file: tex/results.tex
\section{Results} \label{Sec:Results}
To evaluate our LNS solution approach and the benefit of the KOP formulation, we generate globally optimal benchmark solutions of the Dubins Orienteering Problem (DOP) and compare them to exact and heuristic solutions of the KOP. Moreover, we show that the yielded solutions can be used as reference trajectories for multirotors since we demonstrate that they can precisely be tracked by a modern MPC-based flight controller.

\subsection{Benchmark against DOP}
We generate benchmarks on Tsiligirides dataset 2 (see \cite{Tsiligirides.1984}) with slightly modified time budget constraints to be more representative. We chose dataset 2 since it is the only OP dataset whose problem instances could be solved with Gurobi as a commercial solver to optimality within a reasonable time. The kinematic properties we assume for the evaluation are $v_{max} = 3\,\frac{\text{m}}{\text{s}}$ and $a_{max} = 1.5\,\frac{\text{m}}{\text{s}^2}$. Further, we assume that each waypoint can be traversed with eight different and equally distributed heading angles. To determine the optimal solutions for the DOP, we calculate Dubins paths and used their length divided by the constant velocity as edge costs. Dubins paths require a minimum turning radius as input which results from the centripetal force and is calculated by $r = v_{const}^2/a_{max}$. Since $a_{max}$ is a physically given and UAV-specific constant the turning radius can only be changed by a modification of $v_{const}$. However, decreasing the constant velocity to achieve short paths might result in higher flight times. 
	
	To find the best instance dependent constant velocity, we solved the DOP for each constant velocity $v_{const} \in \lbrace 0.1, 0.2, ..., 1.0\rbrace \cdot v_{max}$. The highest collected priorities for the different maximum allowed flight times $C_{max}$ over all $v_{const}$ are presented in the second column of Table \ref{tab:Solutionquality_instance_2}. To directly benchmark the KOP-formulation with the DOP-formulation, we solve the corresponding KOPs with $\numVelo=1$ to optimality by using the associated $v_{const}$ scaled by $\sqrt{2}^{-1}$ as traversal velocities. The scaling is applied to not exceed the maximum allowed velocity of a single axis, which must be bounded to $\left[-3, 3\right]\cdot \frac{1}{\sqrt{2}}\,\left(\frac{\text{m}}{\text{s}}\right)$ to guarantee that the total maximum velocity is not exceeded and hence to be comparable with the DOP. For the same reason, the bound on maximum allowed acceleration for each axis in our trajectory generation is scaled with $\sqrt{2}^{-1}$ as well. The results are given in the third column of Table \ref{tab:Solutionquality_instance_2} ($\text{KOP-1}^\ast$). It can be seen that the optimal solution of the KOP constantly collects approximately $20\%$ more priorities than the best possible solution of the DOP for all instances. Note that the overall highest collected priorities are marked bold in Table \ref{tab:Solutionquality_instance_2}. 
	
	To demonstrate the effectiveness of the proposed LNS, we again conduct the same procedure for the KOP with our heuristic solution approach as solver. For each problem instance and each traversal velocity, we conducted ten runs, each with a different random seed. The overall highest collected priorities and the average collected priorities for the associated traversal velocity in brackets, demonstrating the competitiveness with the exact approach, are given in the fourth column of Table \ref{tab:Solutionquality_instance_2}.
	
	Lastly, we solved the KOP as modeled in Section \ref{Sec:Routing} with multiple options for traversal velocitities. In the first case ($\text{KOP-3}^\text{LNS}$), we consider the set of allowed velocities to be $\setVelo = \lbrace 0, 0.5, 1\rbrace\cdot v_{max}/\sqrt{2}$. In the second case ($\text{KOP-6}^\text{LNS}$), $\setVelo = \lbrace 0, 0.2, 0.4, 0.6, 0.8, 1\rbrace\cdot v_{max}/\sqrt{2}$ holds. Table \ref{tab:Solutionquality_instance_2} shows that for many problem instances the LNS approach yields even better solutions than found by the exact approach ($\text{KOP-1}^\ast$). This is possible since a higher degree of freedom for the travesal velocity is offered, which can successfully be exploited. However, for $C_{max} = 10$\, no improvements compared to the optimal DOP solutions can be found. Sometimes the solution is even worse than the optimal DOP solution. We assume this to be the case since the velocity yielding the best solution for $\text{KOP-1}^\ast$ is not considered. To illustrate the resulting trajectories, an example solution for $\numVelo=6$ and $\numHead=8$ is given in Fig. \ref{fig:kop}. The efficiency of the trajectory is clearly visible since slow velocities only occur when sharp turns are required.	
	
\begin{figure}[]
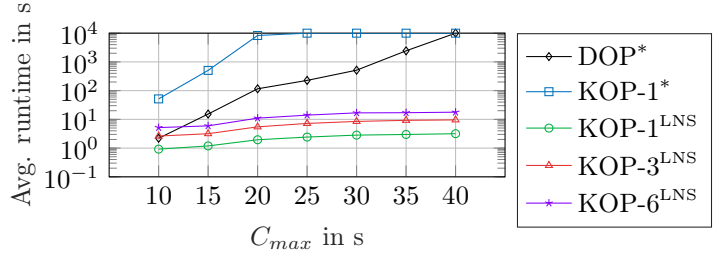

	\centering
	\include{tikz/solution_times}
	\vspace{-8mm}
	\caption{Average runtimes for each solution approach over all problem instances for a specific $C_{max}$.}
	\vspace{-3mm}
	\label{fig:runtimes}
\end{figure}

	Our exact approach, LNS and the trajectory generation are implemented in Python and executed on an Intel Core i7-8565U CPU. The average computation time of a single trajectory is 62\,µs. The average runtimes for each solution method are presented in a logarithmic scale in Figure \ref{fig:runtimes}. Some problem instances for the $\text{DOP}^\ast$ and $\text{KOP-1}^\ast$ could not even be solved to optimality within a runtime of 100.000\,s. In these cases, we set the average runtime for the ten different traversal velocities to be greater than 10.000\,s. Nevertheless, since the dual bounds for the suboptimal solutions provided by Gurobi are worse than the proven optimal solutions for other traversal velocities, the optimality of the values presented in Table \ref{tab:Solutionquality_instance_2} holds. Contrary to the exact approaches, the LNS finds high-quality solutions in a short time and with slowly increasing runtimes for increasing problem sizes. Therefore, we see the benefit of our heuristic solution approach when it comes to larger problem instances.

\begin{table}
	\centering
	\scriptsize
	\begin{tabular}{cccccc}
		\hline
		\hline
		\addlinespace[1ex]
		\multirow{2}{*}{$C_{max}$}& \multicolumn{3}{c}{Best fixed traversal velocity}& \multicolumn{2}{c}{Varying traversal velocity}\\
		\cmidrule(lr){2-4}\cmidrule(lr){5-6}
		& $\text{DOP}^\ast$ & $\text{KOP-1}^\ast$ & $\text{KOP-1}^{\text{\tiny LNS}}$ & $\text{KOP-3}^{\text{\tiny LNS}}$& $\text{KOP-6}^{\text{\tiny LNS}}$\\
		\addlinespace[0.5ex]
		\hline
		\addlinespace[1ex]
		10\,s& 80 &  \textbf{95}   & 80 (80)  & 70 (70)  & 80 (75)\\
		15\,s& 155 & \textbf{180} & \textbf{180} (153) & 175 (170) & 165 (165) \\
		20\,s& 215 & \textbf{250} & 235 (225) & 230 (226.5) & \textbf{250} (237.5) \\
		25\,s& 275 & 325 & 315 (295.2) & \textbf{330} (311.5) & \textbf{330} (316.5) \\
		30\,s& 315 & \textbf{390} & \textbf{390} (367.5) & 385 (369.5) & \textbf{390} (377.5)\\
		35\,s& 370 & 430 & 425 (411.5) & 430 (415) & \textbf{435} (422.5) \\
		40\,s& 415 & \textbf{450} & \textbf{450} (447) & \textbf{450} (446) & \textbf{450} (450) \\
		\hline
		\hline
	\end{tabular}
	\caption{Results for each solution method Tsiligirides dataset 2 containing 21 locations. The asterisk as superscript denotes that the problem instances where solve exactly, whereas LNS indicates the application of our heuristic solver. Average collected priorities for our LNS approach are given in brackets. }
	\label{tab:Solutionquality_instance_2}
\end{table}

\subsection{Trackability of Solutions}
\begin{figure}[t]
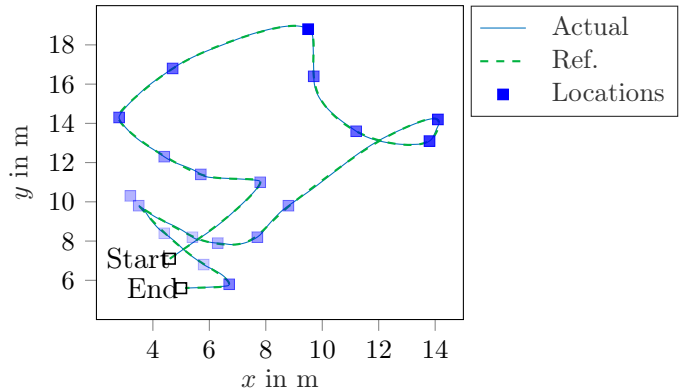

	\centering
	\include{tikz/trajectory_tracking}
	\vspace{-8mm}
	\caption{Plot of the actual and the reference position for the solution of Tsiligirides problem instance number two with $C_{max} = 35$\,s.}
	\vspace{-3mm}
	\label{fig:traj_tracking}
\end{figure}

\begin{figure}[h!]
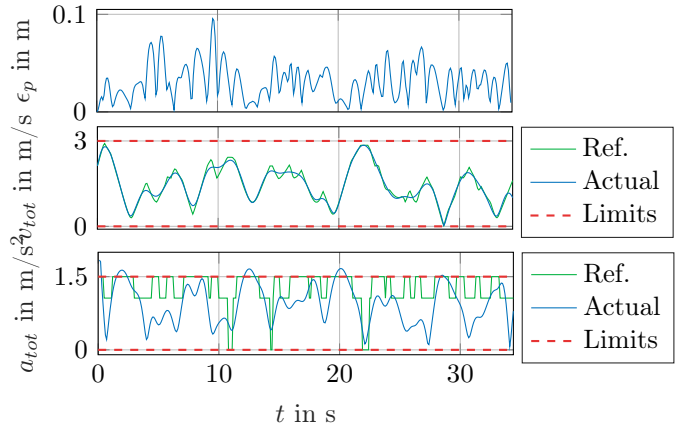

	\centering
	\begin{minipage}[]{0.7\textwidth}
		\begin{flushleft}
			\begin{subfigure}{.5\textwidth}
				\include{tikz/trajectory_tracking_pos_err}
				\vspace{-0.9cm}
			\end{subfigure}
			\begin{subfigure}{.5\textwidth}
				\include{tikz/trajectory_tracking_velo}
				\vspace{-0.9cm}
			\end{subfigure}
			\begin{subfigure}{.5\textwidth}
				\include{tikz/trajectory_tracking_acc}
				\label{fig:traj_tracking_acc}
			\end{subfigure}
		\end{flushleft}
	\end{minipage}
	\vspace{-8mm}
	\caption{Position error $\epsilon_p$, total velocity $v_{tot}$ and total acceleration $a_{tot}$ (with limits) for tracking the reference trajectory shown in Figure \ref{fig:traj_tracking}.}
	\vspace{-3mm}
	\label{fig:err_and_acc}
\end{figure}

We simulatively indicate the precise trackability of the solutions by the utilization of a modern MPC-based UAV trajectory tracking controller. For that, we apply a twelve state dynamic quadrotor model from \cite{Luukkonen.2011} with squared angular velocity for each rotor as the control input to represent real dynamic behavior. Next, we apply a nonlinear MPC according to \cite{Tzorakoleftherakis.2018}, which is provided by the MPC toolbox in MATLAB, with a sampling time of $0.1$\,s and a prediction horizon of $1.8$\,s. We use the position of the trajectory shown in Figure \ref{fig:kop} as a reference for demonstration. Figure \ref{fig:traj_tracking} shows that the reference trajectory is tracked precisely with neglectable overshoots in position due to the deviation between the kinematic model for generating the reference trajectory and the nonlinear dynamic quadrotor model representing real behavior. The tracking error for the position and as well as the actual velocity and acceleration profile are given in blue in Figure \ref{fig:err_and_acc}, whereas the solution of the KOP as reference is given in green. The root mean square error (RMSE) of the obtained trajectory compared with the reference trajectory is $0.035$\,m. It can be seen that the velocity and acceleration profiles of the reference exactly comply with the predefined bounds. The same holds for the actual trajectory with a few minor exceptions for the control inputs, since the applied MPC has constraints on the maximum thrust but not on the maximum acceleration. 

%% file: tikz/solution_times.tex
%
%
\definecolor{mycolor1}{rgb}{0.00000,0.44700,0.74100}
\definecolor{mycolor2}{rgb}{0,0.7,0.25}%
\definecolor{mycolor3}{rgb}{0.9,0.2,0.2}%
\definecolor{mycolor4}{rgb}{0,0,0}%
\definecolor{mycolor5}{rgb}{0.53,0,1}%
\begin{tikzpicture}
	
	\begin{axis}[%
		width=2.05in,
		height=0.75in,
		at={(0.758in,0.481in)},
		scale only axis,
		xmin=5,
		xmax=45,
		xlabel style={font=\color{white!15!black}},
		xticklabel style = {font= \small},
		xlabel={\small$C_{max}$ in s},
		xtick={10, 15, 20, 25, 30, 35, 40},
		ymin=1e-1,
		ymax=1e4,
		ylabel style={font=\color{white!15!black}},
		yticklabel style = {font= \small},
		ymode=log,
		ytick={1e-1,1e0, 10, 100, 1000, 10000},
		ylabel={\small Avg. runtime in s},
		y label style = {at={(0.02,0.5)}},
		axis background/.style={fill=white},
		title style={font=\bfseries},
		xmajorgrids,
		ymajorgrids,
		legend style={legend pos = outer north east ,legend cell align=left, align=left, draw=white!15!black}
		]
		\addplot [color=mycolor4, mark=diamond, mark size=1.5]
		table[row sep=crcr]{%
			10 2.2\\
			15 15.3\\
			20 115.9\\
			25 227\\
			30 514.5\\
			35 2430.5\\
			40 10000\\
		};
		\addlegendentry{\small\text{$\text{DOP}^\ast$}}
	
		\addplot [color=mycolor1, mark=square, mark size=1.5]
		table[row sep=crcr]{%
			10 51.6\\
			15 506.0\\
			20 8306.6\\
			25 10000\\
			30 10000\\
			35 10000\\
			40 10000 \\
		};
		\addlegendentry{\small \text{$\text{KOP-1}^\ast$}}
		
		\addplot [color=mycolor2, mark=o, mark size=1.5]
		table[row sep=crcr]{%
			10 0.92\\
			15 1.19\\
			20 1.95\\
			25 2.42\\
			30 2.83\\
			35 2.98\\
			40 3.17\\
		};
		\addlegendentry{\small \text{$\text{KOP-1}^\text{\tiny LNS}$}}
		
		\addplot [color=mycolor3, mark=triangle, mark size=1.5]
		table[row sep=crcr]{%
			10 2.62\\
			15 3.17\\
			20 5.49\\
			25 7.22\\
			30 8.48\\
			35 9.25\\
			40 9.58\\
		};
		\addlegendentry{\small \text{$\text{KOP-3}^\text{\tiny LNS}$}}
		
		\addplot [color=mycolor5, mark=star, mark size=1.5]
		table[row sep=crcr]{%
			10 5.17\\
			15 5.94\\
			20 11.03\\
			25 14.04\\
			30 16.83\\
			35 17.07\\
			40 17.77\\
		};
		\addlegendentry{\small\text{$\text{KOP-6}^\text{\tiny LNS}$}}
\end{axis}
	
\end{tikzpicture}%

%% file: tikz/trajectory_tracking.tex
%
%
\definecolor{mycolor1}{rgb}{0.00000,0.44700,0.74100}
\definecolor{mycolor2}{rgb}{0,0.7,0.25}%

\begin{tikzpicture}

\begin{axis}[%
	legend cell align={left},
	legend style={
		fill opacity=0.8,
		draw opacity=1,
		text opacity=1,
		at={(0.58,0.3)},
		anchor=north west,
		draw=white!80!black
	},
width=1.9in,
at={(0.758in,0.481in)},
scale only axis,
xmin=2,
xmax=15,
xlabel style={font=\color{white!15!black}},
xticklabel style = {font= \small},
xlabel={\small $x$ in m},
xtick={4,6,8,10,12,14},
ymin=4,
ymax=20,
ylabel style={font=\color{white!15!black}},
yticklabel style = {font= \small},
ylabel=\small {$y$ in m},
y label style = {at={(0.05,0.5)}},
ytick={6,8,10,12,14,16,18},
tick align=outside,
tick pos=left,
axis background/.style={fill=white},
title style={font=\bfseries},
legend style={legend cell align=left, align=left, at={(1.02,1.0)}, draw=white!15!black}
]
\addplot [color=mycolor1]
table[row sep=crcr]{%
	4.6	7.1\\
	4.755303301	7.255303301\\
	4.921213203	7.421213203\\
	5.097729708	7.597729708\\
	5.284852814	7.784852814\\
	5.482582521	7.982136514\\
	5.690795794	8.182093608\\
	5.897010783	8.382050702\\
	6.092619171	8.582007796\\
	6.277620956	8.78196489\\
	6.45201614	8.981921984\\
	6.615804723	9.181879078\\
	6.768986703	9.377444509\\
	6.911562082	9.562446295\\
	7.043530859	9.736841479\\
	7.164893034	9.900630061\\
	7.275648608	10.05381204\\
	7.37579758	10.19638742\\
	7.46533995	10.3283562\\
	7.544275719	10.44971837\\
	7.612604885	10.56047395\\
	7.670327451	10.66062292\\
	7.717443414	10.75016529\\
	7.753952775	10.82910106\\
	7.779855535	10.89743022\\
	7.795151694	10.95515279\\
	7.799980908	11.00252649\\
	7.794041211	11.0390132\\
	7.777494912	11.06489331\\
	7.750342012	11.08016682\\
	7.71258251	11.08992701\\
	7.664216406	11.09968509\\
	7.605243701	11.10944316\\
	7.535664393	11.11920123\\
	7.455478484	11.12895931\\
	7.364685974	11.13871738\\
	7.263286861	11.14847545\\
	7.151281147	11.15823353\\
	7.028668831	11.1679916\\
	6.895449914	11.17774967\\
	6.751624394	11.18750775\\
	6.599153434	11.19726582\\
	6.45384299	11.20702389\\
	6.319139149	11.21678197\\
	6.195041909	11.22654004\\
	6.08155127	11.23648956\\
	5.978667233	11.25356619\\
	5.886389799	11.28124942\\
	5.804718965	11.31953926\\
	5.733654734	11.36843569\\
	5.671277822	11.42872218\\
	5.601095484	11.49550191\\
	5.520306545	11.56249166\\
	5.428911003	11.62948141\\
	5.326908861	11.69647116\\
	5.214300116	11.76346092\\
	5.09108477	11.83045067\\
	4.957262822	11.90233813\\
	4.812834272	11.98482412\\
	4.666012873	12.07791672\\
	4.529645341	12.18161592\\
	4.40388441	12.29592172\\
	4.280042655	12.42008752\\
	4.144567623	12.55273928\\
	3.998485989	12.6857748\\
	3.84586179	12.81881031\\
	3.702304528	12.95184582\\
	3.569353867	13.08488134\\
	3.447009808	13.21791685\\
	3.335272351	13.35095236\\
	3.234141495	13.48398788\\
	3.143617241	13.61569226\\
	3.063699589	13.73811056\\
	2.994388539	13.84992227\\
	2.93568409	13.95112737\\
	2.887586243	14.04172587\\
	2.850094998	14.12171777\\
	2.823210355	14.19110306\\
	2.806932313	14.24988176\\
	2.801260873	14.29805385\\
	2.804867179	14.34561687\\
	2.820331604	14.40352892\\
	2.846402631	14.47204757\\
	2.88308026	14.55117281\\
	2.93036449	14.64090466\\
	2.988255322	14.74124312\\
	3.056752756	14.85218817\\
	3.135856792	14.97373983\\
	3.225567429	15.10589808\\
	3.325884668	15.24866294\\
	3.436808509	15.4020344\\
	3.558338951	15.56601247\\
	3.690475996	15.73792374\\
	3.833219642	15.91028101\\
	3.986569889	16.08263617\\
	4.14909035	16.24947802\\
	4.305240741	16.40571326\\
	4.450784529	16.5513419\\
	4.585721716	16.68636394\\
	4.710480141	16.8094469\\
	4.836706216	16.9271283\\
	4.973538893	17.04468879\\
	5.120978171	17.16224927\\
	5.279024052	17.27980975\\
	5.447676534	17.39737023\\
	5.626935618	17.51493071\\
	5.816801303	17.63249119\\
	6.01727359	17.75005168\\
	6.227516631	17.86761216\\
	6.439688	17.98517264\\
	6.651859369	18.10273312\\
	6.864030738	18.2202936\\
	7.076202107	18.33772001\\
	7.288373476	18.44829074\\
	7.500544845	18.54825487\\
	7.70959634	18.6376124\\
	7.908329145	18.71636332\\
	8.096455348	18.78450765\\
	8.273974949	18.84204537\\
	8.440887948	18.88897649\\
	8.597194346	18.92530101\\
	8.742894142	18.95101893\\
	8.877987336	18.96613024\\
	9.002473929	18.97063496\\
	9.11635392	18.96453307\\
	9.219627309	18.94782458\\
	9.312294096	18.92050949\\
	9.394354282	18.8825878\\
	9.465807866	18.8340595\\
	9.524653054	18.77392895\\
	9.574820735	18.70418148\\
	9.614381813	18.62382741\\
	9.64333629	18.53286674\\
	9.661684165	18.43129947\\
	9.669432699	18.3191256\\
	9.672263115	18.19634512\\
	9.675093532	18.06295804\\
	9.677923949	17.91896436\\
	9.680754366	17.76436408\\
	9.683584782	17.5991572\\
	9.686415199	17.42334372\\
	9.689245616	17.23692363\\
	9.692076032	17.03989694\\
	9.694906449	16.83229675\\
	9.697736866	16.62019808\\
	9.700269402	16.40809941\\
	9.705051797	16.19962551\\
	9.720707141	16.00306397\\
	9.746969087	15.81710903\\
	9.783837635	15.64176069\\
	9.831312784	15.47701895\\
	9.889394535	15.3199954\\
	9.958082888	15.16333589\\
	10.03737784	15.00667639\\
	10.1272794	14.85001688\\
	10.22778756	14.69335737\\
	10.33890232	14.53669787\\
	10.46062368	14.38003836\\
	10.59295164	14.22337885\\
	10.7358862	14.06674387\\
	10.88500107	13.91610891\\
	11.0248385	13.77608055\\
	11.15406934	13.6466588\\
	11.2770716	13.529114\\
	11.40900403	13.4208555\\
	11.55154306	13.32320361\\
	11.70468869	13.23615833\\
	11.86844092	13.15971964\\
	12.04279975	13.09388755\\
	12.22776519	13.03866207\\
	12.41465015	12.99404319\\
	12.59102424	12.96003091\\
	12.75679173	12.93662523\\
	12.91195261	12.92382615\\
	13.0565069	12.92163368\\
	13.19045458	12.92587533\\
	13.31379566	12.93018471\\
	13.42653014	12.93761395\\
	13.52865801	12.95536189\\
	13.62017929	12.98371643\\
	13.70109396	13.02267757\\
	13.77140203	13.07224531\\
	13.83052773	13.13237204\\
	13.87956051	13.20314834\\
	13.91798668	13.28453125\\
	13.95073458	13.37652076\\
	13.98347562	13.47911688\\
	14.01621665	13.59229715\\
	14.04895769	13.70452405\\
	14.08169872	13.80614436\\
	14.11188549	13.89715806\\
	14.13196224	13.97756516\\
	14.14143239	14.04736566\\
	14.14029594	14.10655956\\
	14.12855289	14.15514686\\
	14.10620323	14.19312755\\
	14.07364112	14.21962699\\
	14.03013892	14.2363157\\
	13.97603011	14.24239781\\
	13.91131471	14.23787333\\
	13.8359927	14.22274223\\
	13.75006409	14.19700454\\
	13.65352887	14.16066025\\
	13.54638706	14.11370935\\
	13.42863864	14.05615186\\
	13.30028363	13.98798776\\
	13.16132201	13.90921706\\
	13.01175379	13.81983975\\
	12.85157896	13.71985585\\
	12.68079754	13.60926534\\
	12.49940951	13.48806824\\
	12.3098169	13.35626453\\
	12.11965722	13.21385422\\
	11.92949753	13.0608373\\
	11.73933784	12.89721379\\
	11.54917816	12.72298367\\
	11.35901847	12.53814696\\
	11.16885879	12.34270364\\
	10.9786991	12.13665372\\
	10.78853942	11.92454465\\
	10.59837973	11.71240654\\
	10.40822004	11.50026842\\
	10.21806036	11.28813031\\
	10.02790067	11.0759922\\
	9.837740985	10.86385409\\
	9.647581299	10.65688229\\
	9.457421613	10.4605162\\
	9.271888294	10.27475671\\
	9.096938475	10.09960382\\
	8.932595257	9.935057528\\
	8.778509969	9.780864463\\
	8.635340621	9.624214124\\
	8.502777874	9.456957183\\
	8.380821729	9.280298221\\
	8.269472186	9.109583528\\
	8.168729244	8.949475437\\
	8.078592904	8.799973947\\
	7.999063166	8.661079059\\
	7.930139945	8.532790773\\
	7.865654419	8.415109088\\
	7.792509855	8.308034005\\
	7.70875869	8.211565524\\
	7.614405641	8.12442307\\
	7.513483471	8.04912382\\
	7.412463212	7.984431172\\
	7.311442953	7.930345125\\
	7.210422694	7.88686568\\
	7.109402435	7.853992837\\
	7.008382175	7.831726596\\
	6.907361916	7.820066956\\
	6.806341657	7.819013918\\
	6.705321398	7.828567482\\
	6.604301139	7.848727647\\
	6.50328088	7.875337942\\
	6.403042624	7.892825243\\
	6.311398601	7.899705943\\
	6.222822288	7.903786689\\
	6.123512676	7.918052569\\
	6.013596462	7.94292505\\
	5.894249802	7.976770583\\
	5.780790984	8.011666303\\
	5.677938767	8.046562023\\
	5.585693152	8.081457742\\
	5.504054138	8.119876047\\
	5.433021727	8.168719448\\
	5.371148042	8.227685472\\
	5.300944491	8.297698104\\
	5.220134339	8.378317337\\
	5.128717585	8.468450078\\
	5.026694229	8.560163817\\
	4.914064271	8.651877556\\
	4.790827711	8.743591295\\
	4.65698455	8.835305034\\
	4.512779842	8.927018773\\
	4.371554486	9.018732512\\
	4.240935732	9.110446252\\
	4.12092358	9.202159991\\
	4.011518029	9.29387373\\
	3.91271908	9.385587469\\
	3.824526732	9.473787262\\
	3.746940987	9.551563927\\
	3.679961843	9.618733989\\
	3.623589301	9.67529745\\
	3.57782336	9.72125431\\
	3.542664021	9.756604567\\
	3.518111284	9.781348223\\
	3.504165149	9.795485277\\
	3.500175667	9.799857362\\
	3.507409369	9.792814579\\
	3.525249673	9.775165194\\
	3.553696579	9.746909207\\
	3.592266404	9.708046618\\
	3.633420019	9.658577428\\
	3.674573633	9.598501635\\
	3.715727248	9.527819242\\
	3.756880863	9.446530246\\
	3.798034477	9.354634649\\
	3.843794666	9.25213245\\
	3.900136935	9.139023649\\
	3.967085805	9.015308247\\
	4.044641277	8.880986242\\
	4.13280335	8.736342912\\
	4.231572025	8.594893271\\
	4.340947302	8.464050232\\
	4.460904676	8.342053498\\
	4.582813865	8.211733274\\
	4.704723053	8.070806449\\
	4.826632241	7.919273022\\
	4.94854143	7.757132994\\
	5.070450618	7.594019374\\
	5.192359806	7.441488927\\
	5.314268995	7.299565081\\
	5.436178183	7.168247837\\
	5.555908346	7.047537195\\
	5.665715401	6.937433154\\
	5.764915855	6.837935715\\
	5.85880488	6.747741937\\
	5.96079034	6.66824391\\
	6.073229667	6.590218764\\
	6.183276144	6.512193617\\
	6.282716019	6.43416847\\
	6.371549293	6.356143324\\
	6.449775965	6.278118177\\
	6.517396036	6.20009303\\
	6.574409504	6.122067884\\
	6.620816371	6.044042737\\
	6.656616637	5.967105875\\
	6.6818103	5.89918882\\
	6.696397362	5.841878367\\
	6.699911748	5.795727682\\
	6.693240195	5.759686449\\
	6.675962041	5.734251818\\
	6.648077285	5.719423789\\
	6.609585928	5.711076797\\
	6.560487968	5.702803647\\
	6.500783407	5.694530498\\
	6.430472244	5.686257349\\
	6.34955448	5.677984199\\
	6.258030114	5.66971105\\
	6.155899146	5.661437901\\
	6.043161576	5.653164751\\
	5.919817405	5.644891602\\
	5.785866632	5.636618453\\
	5.641309257	5.628345303\\
	5.48614528	5.620072154\\
	5.320374702	5.611799005\\
	5.144232999	5.603525855\\
	5.144232999	5.603525855\\
	5.144232999	5.603525855\\
	5.144232999	5.603525855\\
	5.144232999	5.603525855\\
};
\addlegendentry{\small Actual}

\addplot [thick, color=mycolor2, dashed]
table[row sep=crcr]{%
	4.6	7.1\\
	4.7501151727439	7.25009692836511\\
	4.90141353382853	7.40120287833579\\
	5.05657121291262	7.55579687644731\\
	5.21844852818598	7.71687368681398\\
	5.38794170939024	7.88598169453562\\
	5.56452463720329	8.06357551272591\\
	5.74667525615571	8.2492513628239\\
	5.93221466718214	8.44191443998923\\
	6.11859435772275	8.63992895519547\\
	6.30313618347395	8.84126243067048\\
	6.48322776185032	9.04363090922254\\
	6.656471847321	9.24464288017038\\
	6.82078760634812	9.44193223456732\\
	6.97446597732784	9.63326643390652\\
	7.11618685879518	9.81662085835362\\
	7.24500757647819	9.99022432838296\\
	7.36033257543633	10.1525873400617\\
	7.46187272218377	10.3025217757866\\
	7.54959836204665	10.4391559542299\\
	7.62368844040981	10.5619449760867\\
	7.68447795771664	10.6706743167122\\
	7.73240643861234	10.7654541735925\\
	7.76797013182724	10.8467026545342\\
	7.79167941060423	10.915116701545\\
	7.80401807908562	10.9716302652966\\
	7.80539988104103	11.0173720312926\\
	7.79612348397564	11.0536235949491\\
	7.77633993226592	11.0817637987819\\
	7.74605209353605	11.1032019186099\\
	7.70515117490375	11.1193059348194\\
	7.65348251683645	11.131333487853\\
	7.59092931299819	11.1403790604489\\
	7.51749979857199	11.1473444414447\\
	7.43340277047177	11.152930754436\\
	7.33912416592074	11.1576499872408\\
	7.23550748066947	11.1618634932336\\
	7.12379713847508	11.1658554766659\\
	7.0056143831163	11.1699236100678\\
	6.88286548729151	11.1744484176745\\
	6.75760340308389	11.1799201421269\\
	6.63186990166285	11.1869284498402\\
	6.50754066164673	11.1961321510364\\
	6.38618432388365	11.2082268621962\\
	6.26893516770448	11.2239094115947\\
	6.15639422127801	11.2438239685806\\
	6.04859779993266	11.2684979559744\\
	5.94508221975603	11.2983037831416\\
	5.84501890854829	11.3334711629759\\
	5.74736227068509	11.374136853054\\
	5.65096470484041	11.4204007883952\\
	5.55464977202895	11.4723667507276\\
	5.45726369568098	11.5301587676048\\
	5.35772004725018	11.593914565752\\
	5.25504580039556	11.6637659171255\\
	5.14843074893678	11.7398171694376\\
	5.03727653405398	11.8221308291193\\
	4.9212380039509	11.9107192965511\\
	4.80025087760125	12.0055356107178\\
	4.67454361246431	12.1064608833186\\
	4.54463384470963	12.2132892872874\\
	4.41131035735502	12.3257113014654\\
	4.27560100503287	12.4432959596608\\
	4.13872754349162	12.5654746843305\\
	4.00205307138215	12.6915203969494\\
	3.8670236701473	12.8205157551751\\
	3.73510550251485	12.9513306058287\\
	3.6077274767106	13.0826366478461\\
	3.48623765601767	13.2129667914659\\
	3.37187399497634	13.3408082163286\\
	3.26574391154504	13.464711442835\\
	3.16880681423305	13.5833993009678\\
	3.08185911481553	13.6958646017405\\
	3.00552709587131	13.8014493797315\\
	2.94027216898683	13.8998996891118\\
	2.88640792932286	13.991390115504\\
	2.844124892582	14.0765177590838\\
	2.81351325091317	14.1562767721184\\
	2.79458290112938	14.2320143195504\\
	2.78728762985317	14.3053655698418\\
	2.79155842197329	14.3781751823926\\
	2.80734430396495	14.4524023440383\\
	2.83464098803556	14.5300087800046\\
	2.87348874908714	14.6128430360705\\
	2.9239333736863	14.7025299367381\\
	2.98595284380829	14.8003600714884\\
	3.05937241060653	14.9071810434564\\
	3.14379859547201	15.0233146115667\\
	3.2385878981722	15.1485304190935\\
	3.3428551914265	15.282090263259\\
	3.45552024751919	15.4228501388971\\
	3.5753840863328	15.5693901410325\\
	3.70122373595738	15.7201428394769\\
	3.83189670240191	15.8735017856841\\
	3.96644101473664	16.0279080666794\\
	4.10415128268644	16.1819230681829\\
	4.24461916446429	16.3342908910245\\
	4.38773654758053	16.4839907054565\\
	4.5336652921958	16.6302770561448\\
	4.68278249228258	16.7726981983415\\
	4.83561645966258	16.9110826529718\\
	4.99278074463511	17.0454945082956\\
	5.15490312840582	17.1761699165747\\
	5.32254910860326	17.3034503790632\\
	5.49614531732158	17.4277206785978\\
	5.67591210279333	17.5493510707222\\
	5.86181419649063	17.6686417306607\\
	6.053534850679	17.785771050914\\
	6.25047414465723	17.9007534845163\\
	6.45176831983793	18.0134136608225\\
	6.65632491678319	18.1233812361422\\
	6.86286840758899	18.2301063567461\\
	7.06999215718673	18.3328914155333\\
	7.27621454778233	18.4309319696373\\
	7.48003245800061	18.5233602852813\\
	7.67996910699693	18.6092872232215\\
	7.8746178951362	18.6878396121785\\
	8.06267992827708	18.7581920369627\\
	8.24299185469571	18.8195912975069\\
	8.41453847454313	18.8713742915201\\
	8.57644911526079	18.912981268281\\
	8.72798673664823	18.9439624065183\\
	8.86854051972655	18.9639762364598\\
	8.99762700044617	18.9727819536346\\
	9.11489961010041	18.9702284371634\\
	9.22016365079414	18.9562420292214\\
	9.31339250420502	18.9308138676932\\
	9.39473972025819	18.8939852798138\\
	9.46454368108598	18.8458278221654\\
	9.52332390876384	18.7864173862524\\
	9.57176463281416	18.7158070576587\\
	9.6106783448332	18.6340049569971\\
	9.64095687285898	18.5409674757544\\
	9.66352956806335	18.4366166009181\\
	9.6793395384526	18.3208788820015\\
	9.68933883660792	18.1937402117062\\
	9.69449853176642	18.0553163584103\\
	9.69582220397435	17.9059307545208\\
	9.69434971398918	17.7461785434128\\
	9.69114913849679	17.5769602530665\\
	9.68730365670558	17.3994760378304\\
	9.68390088878491	17.2151795135779\\
	9.68202687288102	17.0256992474141\\
	9.68276221096758	16.8327415872077\\
	9.68718159685168	16.6379853785942\\
	9.69636304636291	16.4429799642567\\
	9.71139616120268	16.249062977497\\
	9.73336737520485	16.0573127173935\\
	9.76330996579597	15.8685384754745\\
	9.80212646147132	15.6832943351812\\
	9.85050713438461	15.5019038696532\\
	9.90886622171127	15.3245009202392\\
	9.97731043174493	15.1510915703843\\
	10.0556467758829	14.9816284982527\\
	10.143435069902	14.8160815611625\\
	10.2400814098446	14.6544939068989\\
	10.3449437583548	14.4970188540077\\
	10.457418000792	14.3439354378617\\
	10.57699116184	14.1956446933921\\
	10.7032574876584	14.0526561775254\\
	10.83589802288	13.9155737441111\\
	10.9746343049895	13.7850744981237\\
	11.1191715857861	13.6618701149136\\
	11.2691404730285	13.5466552618295\\
	11.4240437949418	13.4400571872186\\
	11.5832176574125	13.3425960090056\\
	11.7458139584441	13.2546599515247\\
	11.910805074169	13.1764955425487\\
	12.0770068910908	13.1082131913905\\
	12.2431104335777	13.0498061022869\\
	12.4077122123095	13.0011757527223\\
	12.5693468688767	12.962165209661\\
	12.7265295853811	12.9326064433608\\
	12.8778112853342	12.9123752984571\\
	13.0218452849679	12.9014335881527\\
	13.1574604471237	12.8998437461636\\
	13.2837294354811	12.9077540349238\\
	13.4000180825832	12.9253580765699\\
	13.5060090756781	12.9528353204311\\
	13.601698232792	12.990279076704\\
	13.6873621784624	13.0376212725351\\
	13.7635016979352	13.0945628503129\\
	13.8307686579615	13.1605207991944\\
	13.8898848433787	13.2346026506047\\
	13.9415629929264	13.3156111509744\\
	13.9864378918253	13.4020765443731\\
	14.0250127264037	13.4923105414876\\
	14.0576235157346	13.5844744549527\\
	14.0844221815143	13.6766540274361\\
	14.1053769815325	13.7669343103196\\
	14.1202875363804	13.8534690860691\\
	14.1288106822366	13.9345404433029\\
	14.1304930918225	14.0086052433913\\
	14.1248068322889	14.0743267936919\\
	14.1111841286537	14.1305912869837\\
	14.0890459206803	14.1765088814406\\
	14.0578188493084	14.2114022497505\\
	14.0169440538326	14.234787433348\\
	13.9658878534187	14.2463511384864\\
	13.9041606109598	14.2459279619409\\
	13.8313442078062	14.2334782063076\\
	13.747124888385	14.2090657516108\\
	13.6513259595459	14.172834413955\\
	13.5439366808568	14.1249802342822\\
	13.425136633077	14.065719999367\\
	13.295309134812	13.9952602059576\\
	13.1550362248808	13.9137731818983\\
	13.0050756817227	13.8213870946858\\
	12.8463250710841	13.718192781536\\
	12.6797800345874	13.6042624653146\\
	12.5064919144466	13.4796710988911\\
	12.327521837597	13.3445179991709\\
	12.143889325298	13.1989531983573\\
	11.9565261294479	13.0432118015866\\
	11.7662461786554	12.8776519791365\\
	11.5737323382314	12.7028016152\\
	11.3795399249739	12.5194181900644\\
	11.184120872022	12.3285344426461\\
	10.9878690981753	12.1314432340185\\
	10.7911775457612	11.9296166364937\\
	10.5944925561477	11.7245995798429\\
	10.3983547502931	11.5179085696232\\
	10.2034211166169	11.3109493876068\\
	10.0104734331507	11.104962038827\\
	9.82042469988828	10.9009968978251\\
	9.63431815785847	10.6999164639217\\
	9.45329992012853	10.5024124747853\\
	9.27855660978958	10.3090280532586\\
	9.11122242431164	10.1201830426243\\
	8.95227354029484	9.93620739787693\\
	8.80243292840332	9.75738093482007\\
	8.66210025897601	9.58397069818793\\
	8.53131342750524	9.41625719533966\\
	8.40974364610363	9.25454780409671\\
	8.29672342034553	9.09918221940931\\
	8.19130347835502	8.95053510360389\\
	8.09233061848908	8.80901801838351\\
	7.99853386432123	8.67508416408768\\
	7.90860120714483	8.54924120793351\\
	7.82123838720159	8.43206122404352\\
	7.73522424913496	8.32416615540116\\
	7.64948067052868	8.22617685831195\\
	7.5631437344709	8.13863299234771\\
	7.47560884279737	8.06191209278518\\
	7.386529255286	7.99617767165512\\
	7.29576632257275	7.94136199869696\\
	7.20331343821355	7.8971735178999\\
	7.10921929806708	7.86312081680006\\
	7.01353275163436	7.8385421467798\\
	6.91628201450708	7.82264007513002\\
	6.8175007124596	7.81452722829052\\
	6.71728892069542	7.81328120416231\\
	6.61586795288102	7.81800379624971\\
	6.5136025660398	7.82788109267472\\
	6.41098899413932	7.8422383752719\\
	6.30861756289328	7.86058015686178\\
	6.20711704900769	7.88261002323466\\
	6.10708587684588	7.90822912331983\\
	6.00900696373835	7.93751346664635\\
	5.913150528669	7.97067093891438\\
	5.8194944251912	8.007985489589\\
	5.7276916875847	8.04976190820022\\
	5.63709440192843	8.09628079204857\\
	5.54682571682438	8.14776825741038\\
	5.45588280041562	8.2043818685877\\
	5.36325052665147	8.26620215974415\\
	5.2680093306567	8.3332174789485\\
	5.16942987365477	8.40530582397476\\
	5.06705193656192	8.48222297235283\\
	4.96074245890817	8.56359880149378\\
	4.85072535649431	8.64893732629005\\
	4.73757980137876	8.73761585000508\\
	4.62220705293772	8.82888522312394\\
	4.50577244210929	8.9218742380593\\
	4.38963395546937	9.0155988240893\\
	4.27526727295945	9.10897799715567\\
	4.16419996060622	9.20085736473158\\
	4.05795918340116	9.29003982850171\\
	3.95802527689649	9.37532021923075\\
	3.86578454565474	9.45551938312659\\
	3.78247963860472	9.52951490764407\\
	3.70915687376146	9.59626661619314\\
	3.64661109632435	9.65483624380787\\
	3.59533830588222	9.70440212298162\\
	3.5555115351265	9.74426815949683\\
	3.52699032708124	9.77386255624938\\
	3.50936219071567	9.79271777335778\\
	3.50200657847929	9.80043426026424\\
	3.50417169232872	9.79665164833315\\
	3.51505585034145	9.78106008364497\\
	3.53388412319269	9.75345526142103\\
	3.55997362429475	9.71380431893967\\
	3.59278445880922	9.66229142516115\\
	3.63194883224389	9.59932109771719\\
	3.67727231695299	9.52547069745235\\
	3.72870827403295	9.4414157839192\\
	3.78631292496923	9.34786238857546\\
	3.85019344519656	9.24550364646802\\
	3.92046382389948	9.13500461777979\\
	3.99721175820537	9.01700984965903\\
	4.0804643097787	8.89216614674673\\
	4.17013903236712	8.76115290968288\\
	4.26598433941088	8.62472230526723\\
	4.3675437248443	8.48374384292711\\
	4.47417278583886	8.33923454657495\\
	4.58509824279691	8.19236424780865\\
	4.69949171281037	8.04443088226853\\
	4.81653439676184	7.89680837240802\\
	4.93545723970609	7.75087761365506\\
	5.05555369038836	7.60795271986922\\
	5.17617256643617	7.46921024062271\\
	5.29670163194635	7.33562121876893\\
	5.4165494380576	7.20789138551252\\
	5.53512894979455	7.0864246568406\\
	5.65184274023911	6.9713208889274\\
	5.76606867122155	6.86240944863361\\
	5.87714702824195	6.75931247703902\\
	5.98437133098164	6.66153027239872\\
	6.08698625381284	6.56853929878448\\
	6.18419641930261	6.47988445741307\\
	6.27518610004425	6.39524694502109\\
	6.35914526611005	6.3144797140781\\
	6.43529666618076	6.23761065519922\\
	6.502920267317	6.16481846452877\\
	6.56137287527819	6.09638897836746\\
	6.61010180601818	6.03266083471903\\
	6.64865234347121	5.97396895594031\\
	6.67666935059057	5.92059302298076\\
	6.69389378651692	5.8727163174823\\
	6.70015464670262	5.83039825092917\\
	6.6953214637055	5.79356984200466\\
	6.67917666770099	5.76199081066094\\
	6.65129189803652	5.73523816524963\\
	6.61106544047227	5.71277687999601\\
	6.55792262903113	5.69403561769737\\
	6.49157119158441	5.67848647568831\\
	6.41223839463226	5.66569427456108\\
	6.32083893441859	5.65529959703087\\
	6.21903519629299	5.64696060060711\\
	6.10917140634141	5.64031758014503\\
	5.99410024752345	5.63500419583763\\
	5.87695355080314	5.63068316156685\\
	5.76090904919409	5.62707685159001\\
	5.64898412413984	5.62397614952673\\
	5.54386589053462	5.62122893702368\\
	5.44777633433438	5.61872237371249\\
	5.36237346225228	5.61637269340281\\
	5.28869646362097	5.6141255137482\\
	5.22717021703114	5.61195979126402\\
	5.17766860637442	5.60988737756117\\
	5.13960710467499	5.60794550508889\\
};
\addlegendentry{\small Ref.}

\node[] at (axis cs:3.5,7.15) {Start};
\node[] at (axis cs:4,5.6) {End};

\addplot [thin, blue, opacity=1.0, mark=square*, mark size=2, mark options={solid}, only marks]
table {%
	9.5 18.8
};
\addlegendentry{\small Locations}

\addplot [thin, blue, opacity=0.4, mark=square*, mark size=2, mark options={solid}, only marks]
table {%
	5.7 11.4
};
\addplot [thin, blue, opacity=0.4, mark=square*, mark size=2, mark options={solid}, only marks]
table {%
	4.4 12.3
};
\addplot [thin, blue, opacity=0.6, mark=square*, mark size=2, mark options={solid}, only marks]
table {%
	2.8 14.3
};
\addplot [thin, blue, opacity=0.3, mark=square*, mark size=2, mark options={solid}, only marks]
table {%
	3.2 10.3
};
\addplot [thin, blue, opacity=0.3, mark=square*, mark size=2, mark options={solid}, only marks]
table {%
	3.5 9.8
};
\addplot [thin, blue, opacity=0.2, mark=square*, mark size=2, mark options={solid}, only marks]
table {%
	4.4 8.4
};
\addplot [thin, blue, opacity=0.4, mark=square*, mark size=2, mark options={solid}, only marks]
table {%
	7.8 11
};
\addplot [thin, blue, opacity=0.4, mark=square*, mark size=2, mark options={solid}, only marks]
table {%
	8.8 9.8
};
\addplot [thin, blue, opacity=0.4, mark=square*, mark size=2, mark options={solid}, only marks]
table {%
	7.7 8.2
};
\addplot [thin, blue, opacity=0.3, mark=square*, mark size=2, mark options={solid}, only marks]
table {%
	6.3 7.9
};
\addplot [thin, blue, opacity=0.2, mark=square*, mark size=2, mark options={solid}, only marks]
table {%
	5.4 8.2
};
\addplot [thin, blue, opacity=0.2, mark=square*, mark size=2, mark options={solid}, only marks]
table {%
	5.8 6.8
};
\addplot [thin, blue, opacity=0.5, mark=square*, mark size=2, mark options={solid}, only marks]
table {%
	6.7 5.8
};
\addplot [thin, blue, opacity=0.8, mark=square*, mark size=2, mark options={solid}, only marks]
table {%
	13.8 13.1
};
\addplot [thin, blue, opacity=0.8, mark=square*, mark size=2, mark options={solid}, only marks]
table {%
	14.1 14.2
};
\addplot [thin, blue, opacity=0.6, mark=square*, mark size=2, mark options={solid}, only marks]
table {%
	11.2 13.6
};
\addplot [thin, blue, opacity=0.6, mark=square*, mark size=2, mark options={solid}, only marks]
table {%
	9.7 16.4
};

\addplot [thin, blue, opacity=0.6, mark=square*, mark size=2, mark options={solid}, only marks]
table {%
	4.7 16.8
};

\addplot [semithick, black, mark=square, mark size=2, mark options={solid}, only marks]
table {%
	4.6	7.1
};

\addplot [semithick, black, mark=square, mark size=2, mark options={solid}, only marks]
table {%
	5 5.6
};

\end{axis}

\end{tikzpicture}%

%% file: tikz/trajectory_tracking_pos_err.tex
%
%
\definecolor{mycolor1}{rgb}{0.00000,0.44700,0.74100}%
\begin{tikzpicture}

\begin{axis}[%
width=2.15in,
height=0.535in,
at={(0.758in,0.481in)},
scale only axis,
xmin=0,
xmax=34.4,
xlabel style={font=\color{white!15!black}},
xticklabel style = {font= \small},
xticklabels= {,,},
ymin=0,
ymax=0.105,
ylabel style={font=\color{white!15!black}},
ylabel={\small $\epsilon_p$ in m},
y label style = {at={(0.05,0.5)}},
yticklabel style = {font= \small},
ytick={0,0.1},
axis background/.style={fill=white},
title style={font=\bfseries},
xmajorgrids,
ymajorgrids,
legend style={legend pos = outer north east ,legend cell align=left, align=left, draw=white!15!black}
]
\addplot [color=mycolor1]
  table[row sep=crcr]{%
	0	0\\
	0.1	0.001800150778486\\
	0.2	0.00702835754410944\\
	0.3	0.0129989320045637\\
	0.4	0.015496533750539\\
	0.5	0.0122534619660003\\
	0.6	0.016978480613331\\
	0.7	0.0286757403733424\\
	0.8	0.0344798286053925\\
	0.9	0.0346455487906327\\
	1	0.0299298495925555\\
	1.1	0.0210020721188636\\
	1.2	0.0114803134585817\\
	1.3	0.00538791002078944\\
	1.4	0.00688808106865751\\
	1.5	0.0116214924264282\\
	1.6	0.0159065644171247\\
	1.7	0.0192193539483131\\
	1.8	0.0214229972385589\\
	1.9	0.0224757396990827\\
	2	0.0224120197473684\\
	2.1	0.0213607920970946\\
	2.2	0.0195629128522027\\
	2.3	0.0173588083269358\\
	2.4	0.0150993131458816\\
	2.5	0.0129325441971275\\
	2.6	0.0105685440409532\\
	2.7	0.00722962526510086\\
	2.8	0.00342120069279193\\
	2.9	0.0100445537507783\\
	3	0.019375275244619\\
	3.1	0.0247627450795626\\
	3.2	0.0273710836847317\\
	3.3	0.0284670617945575\\
	3.4	0.0291834030603658\\
	3.5	0.0301337361197799\\
	3.6	0.0310210273328769\\
	3.7	0.0306292440282534\\
	3.8	0.0272382680142085\\
	3.9	0.0195789181402285\\
	4	0.0139858952702984\\
	4.1	0.0321945202817351\\
	4.2	0.051309664742517\\
	4.3	0.0634562598833518\\
	4.4	0.0695391499993066\\
	4.5	0.0711485644832623\\
	4.6	0.0684993277041034\\
	4.7	0.0594858550408883\\
	4.8	0.0425851668626165\\
	4.9	0.0176418405014246\\
	5	0.0226305567576362\\
	5.1	0.0503695236458521\\
	5.2	0.0687747768682264\\
	5.3	0.0775342076049703\\
	5.4	0.0773580364757013\\
	5.5	0.069146874381711\\
	5.6	0.0544849123530521\\
	5.7	0.037311832558429\\
	5.8	0.0228934836460791\\
	5.9	0.0260689721971648\\
	6	0.0299069814656485\\
	6.1	0.0241413496987296\\
	6.2	0.0173821000301149\\
	6.3	0.0101774879433152\\
	6.4	0.00205031453174446\\
	6.5	0.0178142708954677\\
	6.6	0.0289577019428183\\
	6.7	0.034317377707376\\
	6.8	0.0356089303792551\\
	6.9	0.034851259063564\\
	7	0.0351825245341671\\
	7.1	0.040100538030413\\
	7.2	0.0450405896320318\\
	7.3	0.0480667278007896\\
	7.4	0.0487218327910226\\
	7.5	0.046480513306964\\
	7.6	0.0405540069999393\\
	7.7	0.0298721638137114\\
	7.8	0.0134134182311035\\
	7.9	0.0137695416076096\\
	8	0.037814781025043\\
	8.1	0.0536015917217311\\
	8.2	0.0620862062715635\\
	8.3	0.0651486616118059\\
	8.4	0.064470451406649\\
	8.5	0.061405581714862\\
	8.6	0.056877333862789\\
	8.7	0.0513069528975463\\
	8.8	0.0445752075763992\\
	8.9	0.0360221860558772\\
	9	0.0246131995718172\\
	9.1	0.0110256105549143\\
	9.2	0.0181078514273422\\
	9.3	0.0373505012723954\\
	9.4	0.0601453137146086\\
	9.5	0.0833587012709952\\
	9.6	0.0955571890046253\\
	9.7	0.0935277172576774\\
	9.8	0.0771754996803604\\
	9.9	0.0462468958936174\\
	10	0.0168611729175303\\
	10.1	0.0210856444744466\\
	10.2	0.038394529797959\\
	10.3	0.0512470742234388\\
	10.4	0.0588806896265966\\
	10.5	0.061512073087529\\
	10.6	0.0593381444227988\\
	10.7	0.0524983770984386\\
	10.8	0.04201636925308\\
	10.9	0.0326396970791917\\
	11	0.0232997651669428\\
	11.1	0.012273539897484\\
	11.2	0.00727732404007505\\
	11.3	0.0199614726493168\\
	11.4	0.0314147999558902\\
	11.5	0.040612804200856\\
	11.6	0.0441009923965791\\
	11.7	0.0429915105829605\\
	11.8	0.0386059309484844\\
	11.9	0.032134308059492\\
	12	0.0246061657514126\\
	12.1	0.0168957241446629\\
	12.2	0.00976580361417943\\
	12.3	0.00420675563675711\\
	12.4	0.00370046812734476\\
	12.5	0.00647744057702111\\
	12.6	0.00849652953336008\\
	12.7	0.00954473646699228\\
	12.8	0.00998052841535273\\
	12.9	0.0108942622276186\\
	13	0.0104499674653577\\
	13.1	0.00941134918756147\\
	13.2	0.00722136032223265\\
	13.3	0.00508147381529485\\
	13.4	0.0107033226825228\\
	13.5	0.0182339603652965\\
	13.6	0.0220590759576343\\
	13.7	0.0235627984551409\\
	13.8	0.0242094546446424\\
	13.9	0.0247948307142577\\
	14	0.0248544680190998\\
	14.1	0.0228800209439001\\
	14.2	0.0173783164981238\\
	14.3	0.0119256026137121\\
	14.4	0.0211444893411969\\
	14.5	0.0360422329834878\\
	14.6	0.0510405633884344\\
	14.7	0.0570632315886889\\
	14.8	0.0553655700030683\\
	14.9	0.0466583488571149\\
	15	0.0322527954621892\\
	15.1	0.0195993459729089\\
	15.2	0.0216262950746635\\
	15.3	0.0300378565717138\\
	15.4	0.0369652217549213\\
	15.5	0.0404275548156102\\
	15.6	0.0399468584923005\\
	15.7	0.0359487409240768\\
	15.8	0.0311720146052328\\
	15.9	0.0339760887479778\\
	16	0.0473249693952265\\
	16.1	0.0492358613869308\\
	16.2	0.0364435148011225\\
	16.3	0.0183643646666879\\
	16.4	0.0248629807692211\\
	16.5	0.0377425862456037\\
	16.6	0.0455382948267093\\
	16.7	0.0457845654080885\\
	16.8	0.0370617845588194\\
	16.9	0.0186072391105315\\
	17	0.00954921210370524\\
	17.1	0.0219554150686957\\
	17.2	0.0309113150039634\\
	17.3	0.0364932714514348\\
	17.4	0.0406076441989534\\
	17.5	0.0424500691544394\\
	17.6	0.0378962550142057\\
	17.7	0.029691637156914\\
	17.8	0.0235531135865511\\
	17.9	0.0207469112714498\\
	18	0.0215155525948565\\
	18.1	0.0244459295805912\\
	18.2	0.0278280175222736\\
	18.3	0.0311308974074739\\
	18.4	0.0348772711348336\\
	18.5	0.0378906398598662\\
	18.6	0.0363996859079864\\
	18.7	0.0355854982085337\\
	18.8	0.0412686005919821\\
	18.9	0.0447030760432712\\
	19	0.0458440481044182\\
	19.1	0.0457439830505097\\
	19.2	0.0432216052128753\\
	19.3	0.0384584436913549\\
	19.4	0.0321257324401263\\
	19.5	0.0250059822953311\\
	19.6	0.0177807015048777\\
	19.7	0.0120013823276031\\
	19.8	0.00892983286217859\\
	19.9	0.00942494012932259\\
	20	0.0114027495573415\\
	20.1	0.0129442024910384\\
	20.2	0.0134119074682277\\
	20.3	0.012760197282498\\
	20.4	0.0112355021087228\\
	20.5	0.00924712654107101\\
	20.6	0.00718776403593063\\
	20.7	0.00510292092319374\\
	20.8	0.00255305615504495\\
	20.9	0.00339817926978133\\
	21	0.0112331240115996\\
	21.1	0.0217629345274365\\
	21.2	0.0287325152078838\\
	21.3	0.0322976175568519\\
	21.4	0.0331404258521413\\
	21.5	0.0316611997513345\\
	21.6	0.027841816213789\\
	21.7	0.0211774693884039\\
	21.8	0.0108809598023375\\
	21.9	0.00310480093954807\\
	22	0.0103534999539332\\
	22.1	0.0177068755358982\\
	22.2	0.0244011840981198\\
	22.3	0.0308752713415452\\
	22.4	0.0379289637680866\\
	22.5	0.0420973240563374\\
	22.6	0.0395402851142455\\
	22.7	0.0330277115623126\\
	22.8	0.0240796728526786\\
	22.9	0.0198708277699981\\
	23	0.0336409471579101\\
	23.1	0.0477729459265743\\
	23.2	0.0484468773707429\\
	23.3	0.0368322204818212\\
	23.4	0.0269880929930914\\
	23.5	0.0207274614212793\\
	23.6	0.0157889537592954\\
	23.7	0.0159987122691169\\
	23.8	0.0299420316434469\\
	23.9	0.0501140981207914\\
	24	0.0617096674780054\\
	24.1	0.0627958019552375\\
	24.2	0.0544271969250486\\
	24.3	0.0406340327892746\\
	24.4	0.0285327591422074\\
	24.5	0.0187057542226446\\
	24.6	0.0119376134603242\\
	24.7	0.00938628432903873\\
	24.8	0.0101979280983498\\
	24.9	0.0119382947559419\\
	25	0.0148251136578818\\
	25.1	0.0215095929417793\\
	25.2	0.034184921003189\\
	25.3	0.0495155103672066\\
	25.4	0.0519888117879617\\
	25.5	0.0395996016710434\\
	25.6	0.02611561574172\\
	25.7	0.0188790800432639\\
	25.8	0.00731752401141691\\
	25.9	0.0197429502960426\\
	26	0.0384063991450767\\
	26.1	0.0491542967824407\\
	26.2	0.0526261232076237\\
	26.3	0.0500606175226721\\
	26.4	0.0412711761093784\\
	26.5	0.0384713321105773\\
	26.6	0.0480219425398988\\
	26.7	0.0573762508612315\\
	26.8	0.0634631647799073\\
	26.9	0.066518706476699\\
	27	0.0639340966547526\\
	27.1	0.0540294098805058\\
	27.2	0.0356388113855923\\
	27.3	0.00901140036179012\\
	27.4	0.0189837522338374\\
	27.5	0.0352712890808133\\
	27.6	0.0445204690300419\\
	27.7	0.0480854789303737\\
	27.8	0.0480727745067406\\
	27.9	0.0466896051747277\\
	28	0.0432702920651117\\
	28.1	0.0381463970023603\\
	28.2	0.031977587989053\\
	28.3	0.0253795574923483\\
	28.4	0.0188047347179478\\
	28.5	0.0125136926789006\\
	28.6	0.00658106405668713\\
	28.7	0.00188344760674971\\
	28.8	0.00461628344890146\\
	28.9	0.0120096791844816\\
	29	0.0217197671037718\\
	29.1	0.0342127128021199\\
	29.2	0.0426580433895351\\
	29.3	0.0448094318361847\\
	29.4	0.0408804312465639\\
	29.5	0.0309784070376048\\
	29.6	0.0154530479956932\\
	29.7	0.00740275809233238\\
	29.8	0.0186831838244301\\
	29.9	0.0285438156746836\\
	30	0.0363933673345054\\
	30.1	0.0441989806223068\\
	30.2	0.0451841383444287\\
	30.3	0.0328124401902597\\
	30.4	0.0132356001722178\\
	30.5	0.0194897346265605\\
	30.6	0.0269993326796364\\
	30.7	0.0247768355138449\\
	30.8	0.0144430094280128\\
	30.9	0.0198832819731352\\
	31	0.0314261673141253\\
	31.1	0.0392649260338816\\
	31.2	0.0431769119888524\\
	31.3	0.0427793168642227\\
	31.4	0.0352658938275192\\
	31.5	0.0240236511444813\\
	31.6	0.0235917166321525\\
	31.7	0.0273440253672797\\
	31.8	0.0277130284752347\\
	31.9	0.0329170685454082\\
	32	0.0394773530649965\\
	32.1	0.0429357134665103\\
	32.2	0.0422856243229055\\
	32.3	0.0372772771902731\\
	32.4	0.0278562735370435\\
	32.5	0.0146263116641672\\
	32.6	0.0106486254335487\\
	32.7	0.0227327453684502\\
	32.8	0.031753403667118\\
	32.9	0.0354815527431749\\
	33	0.0347060713515123\\
	33.1	0.0286259228154709\\
	33.2	0.016850024233298\\
	33.3	0.00307162709459065\\
	33.4	0.00777674464283179\\
	33.5	0.0165648954965979\\
	33.6	0.0250947594687299\\
	33.7	0.0340411897071728\\
	33.8	0.0426517046152881\\
	33.9	0.0489032573372268\\
	34	0.0499949022290802\\
	34.1	0.0427789657778451\\
	34.2	0.0240753019631057\\
	34.3	0.0094458891971997\\
};

\end{axis}

\end{tikzpicture}%

%% file: tikz/trajectory_tracking_velo.tex
%
%
\definecolor{mycolor1}{rgb}{0.00000,0.44700,0.74100}
\definecolor{mycolor2}{rgb}{0,0.7,0.25}%
\definecolor{mycolor3}{rgb}{0.9,0.2,0.2}%
\begin{tikzpicture}

\begin{axis}[%
width=2.15in,
height=0.535in,
at={(0.758in,0.481in)},
scale only axis,
xmin=0,
xmax=34.4,
xticklabels= {,,},
xlabel style={font=\color{white!15!black}},
xticklabel style = {font= \small},
ymin=-0.1,
ymax=3.5,
ylabel style={font=\color{white!15!black}},
ylabel={\small $v_{tot}$ in m/$\text{s}$},
y label style = {at={(0.05,0.5)}},
yticklabel style = {font= \small},
ytick={0, 3},
axis background/.style={fill=white},
title style={font=\bfseries},
xmajorgrids,
ymajorgrids,
legend style={legend pos=outer north east, legend cell align=left, align=left, at={(1.02,1.00)}, draw=white!15!black}
]
\addplot [color=mycolor2]
table[row sep=crcr]{%
	0	2.121320344\\
	0.1	2.271320344\\
	0.2	2.421320344\\
	0.3	2.571320344\\
	0.4	2.721320344\\
	0.5	2.849653348\\
	0.6	2.910718579\\
	0.7	2.83457838\\
	0.8	2.760414198\\
	0.9	2.688389572\\
	1	2.61868105\\
	1.1	2.551478471\\
	1.2	2.410064934\\
	1.3	2.261308666\\
	1.4	2.112728279\\
	1.5	1.964363682\\
	1.6	1.816267759\\
	1.7	1.668512053\\
	1.8	1.521195703\\
	1.9	1.37445999\\
	2	1.228512988\\
	2.1	1.083673409\\
	2.2	0.940453058\\
	2.3	0.799722359\\
	2.4	0.663068434\\
	2.5	0.533632501\\
	2.6	0.417948561\\
	2.7	0.331483031\\
	2.8	0.30013497\\
	2.9	0.339530558\\
	3	0.441545581\\
	3.1	0.545492896\\
	3.2	0.650124988\\
	3.3	0.755157268\\
	3.4	0.8604432\\
	3.5	0.96589984\\
	3.6	1.071476785\\
	3.7	1.177141665\\
	3.8	1.282872753\\
	3.9	1.388654925\\
	4	1.494477333\\
	4.1	1.5092952\\
	4.2	1.403467848\\
	4.3	1.297679467\\
	4.4	1.191940435\\
	4.5	1.088260596\\
	4.6	1.00114241\\
	4.7	0.930194133\\
	4.8	0.879338459\\
	4.9	0.852180219\\
	5	0.917528137\\
	5.1	1.009242701\\
	5.2	1.090848317\\
	5.3	1.176362729\\
	5.4	1.264993475\\
	5.5	1.356129681\\
	5.6	1.449298773\\
	5.7	1.591006073\\
	5.8	1.735701231\\
	5.9	1.724887758\\
	6	1.704681137\\
	6.1	1.697538741\\
	6.2	1.840981276\\
	6.3	1.936930222\\
	6.4	2.015337348\\
	6.5	1.996444719\\
	6.6	1.918661141\\
	6.7	1.843699985\\
	6.8	1.771919494\\
	6.9	1.703721735\\
	7	1.639553845\\
	7.1	1.535428913\\
	7.2	1.388640546\\
	7.3	1.242619484\\
	7.4	1.097671991\\
	7.5	0.954287387\\
	7.6	0.813292712\\
	7.7	0.676184629\\
	7.8	0.545899506\\
	7.9	0.42870373\\
	8	0.535810456\\
	8.1	0.665393\\
	8.2	0.802123959\\
	8.3	0.94289864\\
	8.4	1.086145867\\
	8.5	1.23100278\\
	8.6	1.376961454\\
	8.7	1.523705301\\
	8.8	1.671027482\\
	8.9	1.818787466\\
	9	1.966886587\\
	9.1	2.115253612\\
	9.2	2.204469998\\
	9.3	2.272112006\\
	9.4	2.341044632\\
	9.5	2.283900079\\
	9.6	2.133900084\\
	9.7	1.983900091\\
	9.8	1.833900098\\
	9.9	1.697701926\\
	10	1.764098739\\
	10.1	1.844535254\\
	10.2	1.927453305\\
	10.3	2.012546194\\
	10.4	2.099549505\\
	10.5	2.188235377\\
	10.6	2.27840734\\
	10.7	2.369895768\\
	10.8	2.425637166\\
	10.9	2.425637166\\
	11	2.425637166\\
	11.1	2.425637166\\
	11.2	2.417508657\\
	11.3	2.368499982\\
	11.4	2.32330287\\
	11.5	2.206713514\\
	11.6	2.069046315\\
	11.7	1.933215138\\
	11.8	1.799635765\\
	11.9	1.668849006\\
	12	1.541565802\\
	12.1	1.418729446\\
	12.2	1.301599538\\
	12.3	1.191859671\\
	12.4	1.091740625\\
	12.5	1.00412435\\
	12.6	0.932541596\\
	12.7	0.880909726\\
	12.8	0.852859932\\
	12.9	0.850298788\\
	13	0.874381512\\
	13.1	0.922538904\\
	13.2	0.991268394\\
	13.3	1.076637371\\
	13.4	1.175112668\\
	13.5	1.281150461\\
	13.6	1.387192569\\
	13.7	1.493238074\\
	13.8	1.599286299\\
	13.9	1.705336737\\
	14	1.811388999\\
	14.1	1.917442783\\
	14.2	2.02349785\\
	14.3	2.121175535\\
	14.4	2.121175535\\
	14.5	2.12098905\\
	14.6	2.021301066\\
	14.7	1.924031705\\
	14.8	1.833886038\\
	14.9	1.751964041\\
	15	1.679469559\\
	15.1	1.689966451\\
	15.2	1.732540506\\
	15.3	1.780423153\\
	15.4	1.833198463\\
	15.5	1.890456719\\
	15.6	1.951803424\\
	15.7	2.01686554\\
	15.8	2.085295328\\
	15.9	2.151539106\\
	16	2.053950444\\
	16.1	1.903950479\\
	16.2	1.75395052\\
	16.3	1.70091788\\
	16.4	1.715616984\\
	16.5	1.743147049\\
	16.6	1.782913805\\
	16.7	1.834121488\\
	16.8	1.895843252\\
	16.9	1.967089636\\
	17	1.85882728\\
	17.1	1.734630212\\
	17.2	1.614820452\\
	17.3	1.500449336\\
	17.4	1.392857253\\
	17.5	1.287165386\\
	17.6	1.181164167\\
	17.7	1.081495573\\
	17.8	0.995306883\\
	17.9	0.925547897\\
	18	0.876151829\\
	18.1	0.850673348\\
	18.2	0.850830394\\
	18.3	0.877517726\\
	18.4	0.92798834\\
	18.5	1.026541109\\
	18.6	1.127575216\\
	18.7	1.220054302\\
	18.8	1.118241155\\
	18.9	1.017297429\\
	19	0.893891501\\
	19.1	0.765430319\\
	19.2	0.646316588\\
	19.3	0.542739985\\
	19.4	0.465197001\\
	19.5	0.428076872\\
	19.6	0.440772612\\
	19.7	0.501159152\\
	19.8	0.594172109\\
	19.9	0.707050561\\
	20	0.831745451\\
	20.1	0.963680702\\
	20.2	1.100254741\\
	20.3	1.239935682\\
	20.4	1.381781638\\
	20.5	1.525188675\\
	20.6	1.669754621\\
	20.7	1.815202605\\
	20.8	1.961336405\\
	20.9	2.1080134\\
	21	2.255127601\\
	21.1	2.344334412\\
	21.2	2.407904011\\
	21.3	2.474388736\\
	21.4	2.543560006\\
	21.5	2.615204656\\
	21.6	2.689125004\\
	21.7	2.765138545\\
	21.8	2.843077393\\
	21.9	2.84891706\\
	22	2.84891706\\
	22.1	2.84891706\\
	22.2	2.84891706\\
	22.3	2.84891706\\
	22.4	2.84891706\\
	22.5	2.77184507\\
	22.6	2.695657564\\
	22.7	2.550600404\\
	22.8	2.400600429\\
	22.9	2.250600458\\
	23	2.12014277\\
	23.1	2.126876316\\
	23.2	2.144108291\\
	23.3	2.111639654\\
	23.4	1.964859645\\
	23.5	1.818605186\\
	23.6	1.673014112\\
	23.7	1.528276027\\
	23.8	1.384853431\\
	23.9	1.312681915\\
	24	1.284973272\\
	24.1	1.274432741\\
	24.2	1.281112866\\
	24.3	1.229004706\\
	24.4	1.17184416\\
	24.5	1.121822136\\
	24.6	1.07993105\\
	24.7	1.047147193\\
	24.8	1.024345358\\
	24.9	1.012200366\\
	25	1.011096311\\
	25.1	1.021069008\\
	25.2	1.041800407\\
	25.3	1.035591828\\
	25.4	0.977099459\\
	25.5	0.863551303\\
	25.6	0.944325914\\
	25.7	1.064275074\\
	25.8	1.191054756\\
	25.9	1.237826816\\
	26	1.136456214\\
	26.1	1.036026171\\
	26.2	0.936839221\\
	26.3	0.878799256\\
	26.4	0.8515974\\
	26.5	0.916479132\\
	26.6	1.066478992\\
	26.7	1.216478887\\
	26.8	1.332898309\\
	26.9	1.411751438\\
	27	1.493976687\\
	27.1	1.579047359\\
	27.2	1.666527763\\
	27.3	1.72864274\\
	27.4	1.63970165\\
	27.5	1.552912515\\
	27.6	1.46865689\\
	27.7	1.387396425\\
	27.8	1.309688749\\
	27.9	1.173578914\\
	28	1.023579028\\
	28.1	0.873579181\\
	28.2	0.723579397\\
	28.3	0.573579726\\
	28.4	0.423580289\\
	28.5	0.273581468\\
	28.6	0.123585511\\
	28.7	0.025985092\\
	28.8	0.175955179\\
	28.9	0.325952796\\
	29	0.475951915\\
	29.1	0.60367589\\
	29.2	0.685101876\\
	29.3	0.772531281\\
	29.4	0.864143842\\
	29.5	0.958741143\\
	29.6	1.055520999\\
	29.7	1.192769969\\
	29.8	1.33497573\\
	29.9	1.478722489\\
	30	1.623600998\\
	30.1	1.739890859\\
	30.2	1.713674473\\
	30.3	1.70033532\\
	30.4	1.746169555\\
	30.5	1.823611542\\
	30.6	1.903814591\\
	30.7	1.986444295\\
	30.8	2.071210245\\
	30.9	1.994325508\\
	31	1.911483269\\
	31.1	1.831038598\\
	31.2	1.753321535\\
	31.3	1.630007628\\
	31.4	1.480007765\\
	31.5	1.330007933\\
	31.6	1.278236245\\
	31.7	1.326604753\\
	31.8	1.392605286\\
	31.9	1.306103231\\
	32	1.222686313\\
	32.1	1.143030179\\
	32.2	1.067976666\\
	32.3	0.998564133\\
	32.4	0.936048341\\
	32.5	0.881897197\\
	32.6	0.793176234\\
	32.7	0.656970986\\
	32.8	0.528292735\\
	32.9	0.413671677\\
	33	0.329881428\\
	33.1	0.302521497\\
	33.2	0.345277796\\
	33.3	0.445692395\\
	33.4	0.5502674\\
	33.5	0.655321838\\
	33.6	0.760657092\\
	33.7	0.866170717\\
	33.8	0.971804616\\
	33.9	1.077523416\\
	34	1.183304361\\
	34.1	1.289132154\\
	34.2	1.394996133\\
	34.3	1.500888641\\
	34.4	1.606804036\\
};
\addlegendentry{\small Ref.}

\addplot [color=mycolor1]
  table[row sep=crcr]{%
	0	2.12132034355964\\
	0.1	2.30528149759884\\
	0.2	2.48584031607775\\
	0.3	2.61831102445004\\
	0.4	2.71264162424302\\
	0.5	2.77142723684162\\
	0.6	2.7970156578115\\
	0.7	2.79203131492481\\
	0.8	2.75930753763284\\
	0.9	2.70183270785276\\
	1	2.62268299522221\\
	1.1	2.5249465255161\\
	1.2	2.41164852269305\\
	1.3	2.2856496036901\\
	1.4	2.14953578268549\\
	1.5	2.00554710260137\\
	1.6	1.85556946498228\\
	1.7	1.70122057855745\\
	1.8	1.54398402788997\\
	1.9	1.3853457294035\\
	2	1.2269092734545\\
	2.1	1.07049554979596\\
	2.2	0.918260950039441\\
	2.3	0.772906188973477\\
	2.4	0.638106794173067\\
	2.5	0.519387083558013\\
	2.6	0.425893989581397\\
	2.7	0.371516380467687\\
	2.8	0.368786793647426\\
	2.9	0.415120409462246\\
	3	0.494466210475737\\
	3.1	0.591595224017525\\
	3.2	0.696729167784954\\
	3.3	0.803658351881795\\
	3.4	0.907837671538591\\
	3.5	1.00508937896883\\
	3.6	1.09122094547731\\
	3.7	1.16234088920323\\
	3.8	1.21540160616856\\
	3.9	1.24866054182176\\
	4	1.26192019820934\\
	4.1	1.25656858536005\\
	4.2	1.23547906010543\\
	4.3	1.20289079590911\\
	4.4	1.16423904646729\\
	4.5	1.12557818128926\\
	4.6	1.09253727696182\\
	4.7	1.06929672757021\\
	4.8	1.05833076993137\\
	4.9	1.06070458604505\\
	5	1.07662120306794\\
	5.1	1.10575849817079\\
	5.2	1.14738673945327\\
	5.3	1.20036159026653\\
	5.4	1.26308241125733\\
	5.5	1.33349581018198\\
	5.6	1.4091880386114\\
	5.7	1.48749170512506\\
	5.8	1.56557468092033\\
	5.9	1.64052494278409\\
	6	1.70943284004612\\
	6.1	1.76947764077097\\
	6.2	1.81803450578961\\
	6.3	1.8527843991166\\
	6.4	1.87164765159869\\
	6.5	1.87277688057937\\
	6.6	1.85475279462014\\
	6.7	1.81681473704081\\
	6.8	1.75902960391597\\
	6.9	1.68240736857622\\
	7	1.58896686507223\\
	7.1	1.4817303918512\\
	7.2	1.3646334379605\\
	7.3	1.24236102462568\\
	7.4	1.12016126312377\\
	7.5	1.00370934653893\\
	7.6	0.899144645835868\\
	7.7	0.813297865428086\\
	7.8	0.753380329220547\\
	7.9	0.725858291621749\\
	8	0.734509706791415\\
	8.1	0.778916001881921\\
	8.2	0.854955789786782\\
	8.3	0.956653127056177\\
	8.4	1.07760970955464\\
	8.5	1.21133950007588\\
	8.6	1.35120988050497\\
	8.7	1.49056600934022\\
	8.8	1.62307599049783\\
	8.9	1.74320248563297\\
	9	1.84667487938095\\
	9.1	1.93079848940184\\
	9.2	1.99458017278952\\
	9.3	2.03869695764622\\
	9.4	2.0652973482625\\
	9.5	2.07763947852683\\
	9.6	2.07966989481445\\
	9.7	2.07566894936711\\
	9.8	2.06990574312159\\
	9.9	2.06626801898422\\
	10	2.06792807851914\\
	10.1	2.07705214063112\\
	10.2	2.09461411584335\\
	10.3	2.12034549174613\\
	10.4	2.15281541536453\\
	10.5	2.18962235904734\\
	10.6	2.22767056193489\\
	10.7	2.26349176707288\\
	10.8	2.29356704734918\\
	10.9	2.31460989152506\\
	11	2.32378797488288\\
	11.1	2.31887801648763\\
	11.2	2.29836527825278\\
	11.3	2.26148099878586\\
	11.4	2.20815160100757\\
	11.5	2.13895274447564\\
	11.6	2.05505401349309\\
	11.7	1.95814132665444\\
	11.8	1.85029184841995\\
	11.9	1.73382434102059\\
	12	1.611199625295\\
	12.1	1.48503218179475\\
	12.2	1.35823142797668\\
	12.3	1.23423158018239\\
	12.4	1.1172662708904\\
	12.5	1.01260577027338\\
	12.6	0.926558995383798\\
	12.7	0.865862214188862\\
	12.8	0.836181528688835\\
	12.9	0.840039732715288\\
	13	0.875613088632247\\
	13.1	0.937666765077635\\
	13.2	1.01975736805534\\
	13.3	1.1158752034051\\
	13.4	1.22094871726828\\
	13.5	1.33063629049081\\
	13.6	1.4408648195255\\
	13.7	1.54760373744342\\
	13.8	1.64694773268802\\
	13.9	1.73537060145284\\
	14	1.81001917085611\\
	14.1	1.86895162504268\\
	14.2	1.91127230091183\\
	14.3	1.93718359682102\\
	14.4	1.94802754906708\\
	14.5	1.94622966669611\\
	14.6	1.93512509010161\\
	14.7	1.91860441053525\\
	14.8	1.90060497340869\\
	14.9	1.88462283852072\\
	15	1.8732578420184\\
	15.1	1.86787294120301\\
	15.2	1.86848410989152\\
	15.3	1.87392624525097\\
	15.4	1.88227964598703\\
	15.5	1.89143996378466\\
	15.6	1.89962412013136\\
	15.7	1.90566618448728\\
	15.8	1.90904384236622\\
	15.9	1.90966456313678\\
	16	1.90758558259972\\
	16.1	1.90287951405229\\
	16.2	1.89559703002175\\
	16.3	1.88568439709937\\
	16.4	1.87288029301533\\
	16.5	1.85664744169532\\
	16.6	1.83616689056525\\
	16.7	1.8103959124262\\
	16.8	1.77817465041787\\
	16.9	1.73832018175325\\
	17	1.68974252064916\\
	17.1	1.63161832224566\\
	17.2	1.56361694014056\\
	17.3	1.48620903876571\\
	17.4	1.40097013839127\\
	17.5	1.31077320897546\\
	17.6	1.21976721381811\\
	17.7	1.13315013857132\\
	17.8	1.05672552544078\\
	17.9	0.996094044247911\\
	18	0.955354973825019\\
	18.1	0.935623991512523\\
	18.2	0.934178170536007\\
	18.3	0.945022926581362\\
	18.4	0.960604463983106\\
	18.5	0.97360175397029\\
	18.6	0.978044901146165\\
	18.7	0.969718694379871\\
	18.8	0.946148527311562\\
	18.9	0.906443410210286\\
	19	0.85115589818569\\
	19.1	0.782250496084764\\
	19.2	0.703281156374139\\
	19.3	0.619948632068638\\
	19.4	0.541257863252906\\
	19.5	0.481081214156193\\
	19.6	0.457501851831196\\
	19.7	0.484144900443984\\
	19.8	0.55918561254898\\
	19.9	0.669470143093756\\
	20	0.801725344824551\\
	20.1	0.946550823227398\\
	20.2	1.09770574218919\\
	20.3	1.25087402540089\\
	20.4	1.40288504227364\\
	20.5	1.55136440484342\\
	20.6	1.69456393002346\\
	20.7	1.83124443597609\\
	20.8	1.96057322867263\\
	20.9	2.08203585502361\\
	21	2.19539274282449\\
	21.1	2.3006614549591\\
	21.2	2.39803410362481\\
	21.3	2.48769749015777\\
	21.4	2.56967489405453\\
	21.5	2.64356021312284\\
	21.6	2.70834587961452\\
	21.7	2.76255812489385\\
	21.8	2.80468185157546\\
	21.9	2.83344355726811\\
	22	2.84798111162785\\
	22.1	2.84794694675994\\
	22.2	2.83349808060859\\
	22.3	2.8051020892241\\
	22.4	2.76329389321556\\
	22.5	2.70861054011049\\
	22.6	2.64170376117053\\
	22.7	2.56356627828943\\
	22.8	2.47574179105099\\
	22.9	2.38034220426893\\
	23	2.27983724052939\\
	23.1	2.17672835961603\\
	23.2	2.07323703657559\\
	23.3	1.97109155399922\\
	23.4	1.87144336747822\\
	23.5	1.77490944904634\\
	23.6	1.68169174913542\\
	23.7	1.59168553828194\\
	23.8	1.50466257881055\\
	23.9	1.4205900259909\\
	24	1.3398980499059\\
	24.1	1.26360374406344\\
	24.2	1.193274587883\\
	24.3	1.13073940032531\\
	24.4	1.07777280575874\\
	24.5	1.03584825358925\\
	24.6	1.0058906231442\\
	24.7	0.988062449612595\\
	24.8	0.981638410894852\\
	24.9	0.984947253780286\\
	25	0.995347649626816\\
	25.1	1.00953546411419\\
	25.2	1.02410471196983\\
	25.3	1.03611164430522\\
	25.4	1.0434814616636\\
	25.5	1.04521009357424\\
	25.6	1.04139128927683\\
	25.7	1.03317385324521\\
	25.8	1.02279657206915\\
	25.9	1.0134652923882\\
	26	1.00886384161576\\
	26.1	1.01241872091246\\
	26.2	1.0265840473195\\
	26.3	1.05243312264583\\
	26.4	1.08965521343989\\
	26.5	1.13675793345038\\
	26.6	1.19130815451883\\
	26.7	1.25017642040508\\
	26.8	1.30979161153972\\
	26.9	1.36639455739411\\
	27	1.41627260019866\\
	27.1	1.4559464360414\\
	27.2	1.48238175248832\\
	27.3	1.4931182667022\\
	27.4	1.48635313045843\\
	27.5	1.46095057334648\\
	27.6	1.41636641982955\\
	27.7	1.35261810622667\\
	27.8	1.27029402334372\\
	27.9	1.17055625695591\\
	28	1.05512761766352\\
	28.1	0.926270973301916\\
	28.2	0.786748774858792\\
	28.3	0.639674719466206\\
	28.4	0.488244408194086\\
	28.5	0.335340074544696\\
	28.6	0.183213837471762\\
	28.7	0.0341380008626771\\
	28.8	0.115264730090791\\
	28.9	0.259866818507163\\
	29	0.401573780682437\\
	29.1	0.539689111303022\\
	29.2	0.673351539554064\\
	29.3	0.801746602774064\\
	29.4	0.924446227956698\\
	29.5	1.04145462830113\\
	29.6	1.15302708437737\\
	29.7	1.25943942955449\\
	29.8	1.36082778718062\\
	29.9	1.45706943945304\\
	30	1.54763057895809\\
	30.1	1.63140706206456\\
	30.2	1.70657855423344\\
	30.3	1.77093385729362\\
	30.4	1.82238108919368\\
	30.5	1.8593363374511\\
	30.6	1.88094847721684\\
	30.7	1.88715949137353\\
	30.8	1.87863764795006\\
	30.9	1.85664105846499\\
	31	1.82287685717635\\
	31.1	1.77939152635636\\
	31.2	1.7283920253275\\
	31.3	1.67198819073609\\
	31.4	1.61192368570698\\
	31.5	1.54939502721659\\
	31.6	1.48498331671162\\
	31.7	1.41865939874937\\
	31.8	1.34988946010171\\
	31.9	1.27786013463584\\
	32	1.20172936645212\\
	32.1	1.12082697239164\\
	32.2	1.03480086037264\\
	32.3	0.943733154909372\\
	32.4	0.848254582447851\\
	32.5	0.749688739885441\\
	32.6	0.650288404273145\\
	32.7	0.55368566819855\\
	32.8	0.465759847061435\\
	32.9	0.395961481814415\\
	33	0.357621760704699\\
	33.1	0.364315821577321\\
	33.2	0.417928948266715\\
	33.3	0.507605758121822\\
	33.4	0.61949737081737\\
	33.5	0.741377437350975\\
	33.6	0.862374335592292\\
	33.7	0.972588866319616\\
	33.8	1.06352033683417\\
	33.9	1.12872832565891\\
	34	1.1642288378779\\
	34.1	1.16846492686463\\
	34.2	1.14201334360382\\
	34.3	1.08724776253163\\
	34.4	1.00804400425644\\
};
\addlegendentry{\small Actual}

\addplot [thick, color=mycolor3, dashed]
table[row sep=crcr]{%
	0	3\\
	35	3\\
};
\addlegendentry{\small Limits}

\addplot [thick, color=mycolor3, dashed, forget plot]
table[row sep=crcr]{%
	0	0\\
	35	0\\
};
\end{axis}

\end{tikzpicture}%

%% file: tikz/trajectory_tracking_acc.tex
%
%
\definecolor{mycolor1}{rgb}{0.00000,0.44700,0.74100}
\definecolor{mycolor2}{rgb}{0,0.7,0.25}%
\definecolor{mycolor3}{rgb}{0.9,0.2,0.2}%
\begin{tikzpicture}

\begin{axis}[
	width=2.15in,
	height=0.535in,
	at={(0.758in,0.481in)},
	scale only axis,
	xmin=0,
	xmax=34.4,
	xlabel style={font=\color{white!15!black}},
	xlabel={\small$t$ in s},
	xticklabel style = {font= \small},
	ymin=-0.1,
	ymax=2,
	ylabel style={font=\color{white!15!black}},
	yticklabel style = {font= \small},
	ylabel={\small $a_{tot}$ in m/$\text{s}^2$},
	y label style = {at={(0.05,0.5)}},
	ytick={0, 1.5},
	axis background/.style={fill=white},
	title style={font=\bfseries},
	xmajorgrids,
	ymajorgrids,
	legend style={legend pos=outer north east, legend cell align=left, align=left, at={(1.02,1.00)}, draw=white!15!black}
	]
	\addplot [color=mycolor2]
	table[row sep=crcr]{%
		0	1.5\\
		0.1	1.5\\
		0.2	1.5\\
		0.3	1.5\\
		0.4	1.5\\
		0.5	1.060660172\\
		0.6	1.060660172\\
		0.7	1.060660172\\
		0.8	1.060660172\\
		0.9	1.060660172\\
		1	1.060660172\\
		1.1	1.060660172\\
		1.2	1.5\\
		1.3	1.5\\
		1.4	1.5\\
		1.5	1.5\\
		1.6	1.5\\
		1.7	1.5\\
		1.8	1.5\\
		1.9	1.5\\
		2	1.5\\
		2.1	1.5\\
		2.2	1.5\\
		2.3	1.5\\
		2.4	1.5\\
		2.5	1.5\\
		2.6	1.5\\
		2.7	1.5\\
		2.8	1.5\\
		2.9	1.5\\
		3	1.060660172\\
		3.1	1.060660172\\
		3.2	1.060660172\\
		3.3	1.060660172\\
		3.4	1.060660172\\
		3.5	1.060660172\\
		3.6	1.060660172\\
		3.7	1.060660172\\
		3.8	1.060660172\\
		3.9	1.060660172\\
		4	1.060660172\\
		4.1	1.060660172\\
		4.2	1.060660172\\
		4.3	1.060660172\\
		4.4	1.060660172\\
		4.5	1.5\\
		4.6	1.5\\
		4.7	1.5\\
		4.8	1.5\\
		4.9	1.5\\
		5	1.5\\
		5.1	1.060660172\\
		5.2	1.060660172\\
		5.3	1.060660172\\
		5.4	1.060660172\\
		5.5	1.060660172\\
		5.6	1.060660172\\
		5.7	1.5\\
		5.8	1.5\\
		5.9	1.5\\
		6	1.5\\
		6.1	1.5\\
		6.2	1.5\\
		6.3	1.060660172\\
		6.4	1.060660172\\
		6.5	1.060660172\\
		6.6	1.060660172\\
		6.7	1.060660172\\
		6.8	1.060660172\\
		6.9	1.060660172\\
		7	1.060660172\\
		7.1	1.5\\
		7.2	1.5\\
		7.3	1.5\\
		7.4	1.5\\
		7.5	1.5\\
		7.6	1.5\\
		7.7	1.5\\
		7.8	1.5\\
		7.9	1.5\\
		8	1.5\\
		8.1	1.5\\
		8.2	1.5\\
		8.3	1.5\\
		8.4	1.5\\
		8.5	1.5\\
		8.6	1.5\\
		8.7	1.5\\
		8.8	1.5\\
		8.9	1.5\\
		9	1.5\\
		9.1	1.5\\
		9.2	1.060660172\\
		9.3	1.060660172\\
		9.4	1.5\\
		9.5	1.5\\
		9.6	1.5\\
		9.7	1.5\\
		9.8	1.5\\
		9.9	1.5\\
		10	1.060660172\\
		10.1	1.060660172\\
		10.2	1.060660172\\
		10.3	1.060660172\\
		10.4	1.060660172\\
		10.5	1.060660172\\
		10.6	1.060660172\\
		10.7	1.060660172\\
		10.8	0\\
		10.9	0\\
		11	0\\
		11.1	0\\
		11.2	1.060660172\\
		11.3	1.060660172\\
		11.4	1.060660172\\
		11.5	1.5\\
		11.6	1.5\\
		11.7	1.5\\
		11.8	1.5\\
		11.9	1.5\\
		12	1.5\\
		12.1	1.5\\
		12.2	1.5\\
		12.3	1.5\\
		12.4	1.5\\
		12.5	1.5\\
		12.6	1.5\\
		12.7	1.5\\
		12.8	1.5\\
		12.9	1.5\\
		13	1.5\\
		13.1	1.5\\
		13.2	1.5\\
		13.3	1.5\\
		13.4	1.060660172\\
		13.5	1.060660172\\
		13.6	1.060660172\\
		13.7	1.060660172\\
		13.8	1.060660172\\
		13.9	1.060660172\\
		14	1.060660172\\
		14.1	1.060660172\\
		14.2	1.060660172\\
		14.3	0\\
		14.4	0\\
		14.5	1.060660172\\
		14.6	1.5\\
		14.7	1.5\\
		14.8	1.5\\
		14.9	1.5\\
		15	1.5\\
		15.1	1.060660172\\
		15.2	1.060660172\\
		15.3	1.060660172\\
		15.4	1.060660172\\
		15.5	1.060660172\\
		15.6	1.060660172\\
		15.7	1.060660172\\
		15.8	1.060660172\\
		15.9	1.5\\
		16	1.5\\
		16.1	1.5\\
		16.2	1.5\\
		16.3	1.5\\
		16.4	1.5\\
		16.5	1.5\\
		16.6	1.5\\
		16.7	1.5\\
		16.8	1.5\\
		16.9	1.5\\
		17	1.5\\
		17.1	1.5\\
		17.2	1.5\\
		17.3	1.5\\
		17.4	1.5\\
		17.5	1.060660172\\
		17.6	1.060660172\\
		17.7	1.5\\
		17.8	1.5\\
		17.9	1.5\\
		18	1.5\\
		18.1	1.5\\
		18.2	1.5\\
		18.3	1.5\\
		18.4	1.5\\
		18.5	1.060660172\\
		18.6	1.060660172\\
		18.7	1.060660172\\
		18.8	1.060660172\\
		18.9	1.060660172\\
		19	1.5\\
		19.1	1.5\\
		19.2	1.5\\
		19.3	1.5\\
		19.4	1.5\\
		19.5	1.5\\
		19.6	1.5\\
		19.7	1.5\\
		19.8	1.5\\
		19.9	1.5\\
		20	1.5\\
		20.1	1.5\\
		20.2	1.5\\
		20.3	1.5\\
		20.4	1.5\\
		20.5	1.5\\
		20.6	1.5\\
		20.7	1.5\\
		20.8	1.5\\
		20.9	1.5\\
		21	1.5\\
		21.1	1.060660172\\
		21.2	1.060660172\\
		21.3	1.060660172\\
		21.4	1.060660172\\
		21.5	1.060660172\\
		21.6	1.060660172\\
		21.7	1.060660172\\
		21.8	1.060660172\\
		21.9	0\\
		22	0\\
		22.1	0\\
		22.2	0\\
		22.3	0\\
		22.4	0\\
		22.5	1.060660172\\
		22.6	1.060660172\\
		22.7	1.5\\
		22.8	1.5\\
		22.9	1.5\\
		23	1.5\\
		23.1	1.5\\
		23.2	1.5\\
		23.3	1.5\\
		23.4	1.5\\
		23.5	1.5\\
		23.6	1.5\\
		23.7	1.5\\
		23.8	1.060660172\\
		23.9	1.5\\
		24	1.5\\
		24.1	1.5\\
		24.2	1.5\\
		24.3	1.060660172\\
		24.4	1.060660172\\
		24.5	1.060660172\\
		24.6	1.060660172\\
		24.7	1.060660172\\
		24.8	1.060660172\\
		24.9	1.060660172\\
		25	1.060660172\\
		25.1	1.060660172\\
		25.2	1.060660172\\
		25.3	1.060660172\\
		25.4	1.5\\
		25.5	1.5\\
		25.6	1.5\\
		25.7	1.5\\
		25.8	1.5\\
		25.9	1.060660172\\
		26	1.060660172\\
		26.1	1.060660172\\
		26.2	1.060660172\\
		26.3	1.5\\
		26.4	1.5\\
		26.5	1.5\\
		26.6	1.5\\
		26.7	1.5\\
		26.8	1.060660172\\
		26.9	1.060660172\\
		27	1.060660172\\
		27.1	1.060660172\\
		27.2	1.060660172\\
		27.3	1.060660172\\
		27.4	1.060660172\\
		27.5	1.060660172\\
		27.6	1.060660172\\
		27.7	1.060660172\\
		27.8	1.060660172\\
		27.9	1.5\\
		28	1.5\\
		28.1	1.5\\
		28.2	1.5\\
		28.3	1.5\\
		28.4	1.5\\
		28.5	1.5\\
		28.6	1.5\\
		28.7	1.5\\
		28.8	1.5\\
		28.9	1.5\\
		29	1.5\\
		29.1	1.060660172\\
		29.2	1.060660172\\
		29.3	1.060660172\\
		29.4	1.060660172\\
		29.5	1.060660172\\
		29.6	1.060660172\\
		29.7	1.5\\
		29.8	1.5\\
		29.9	1.5\\
		30	1.5\\
		30.1	1.5\\
		30.2	1.5\\
		30.3	1.5\\
		30.4	1.060660172\\
		30.5	1.060660172\\
		30.6	1.060660172\\
		30.7	1.060660172\\
		30.8	1.060660172\\
		30.9	1.060660172\\
		31	1.060660172\\
		31.1	1.060660172\\
		31.2	1.060660172\\
		31.3	1.5\\
		31.4	1.5\\
		31.5	1.5\\
		31.6	1.5\\
		31.7	1.060660172\\
		31.8	1.060660172\\
		31.9	1.060660172\\
		32	1.060660172\\
		32.1	1.060660172\\
		32.2	1.060660172\\
		32.3	1.060660172\\
		32.4	1.060660172\\
		32.5	1.060660172\\
		32.6	1.5\\
		32.7	1.5\\
		32.8	1.5\\
		32.9	1.5\\
		33	1.5\\
		33.1	1.5\\
		33.2	1.5\\
		33.3	1.060660172\\
		33.4	1.060660172\\
		33.5	1.060660172\\
		33.6	1.060660172\\
		33.7	1.060660172\\
		33.8	1.060660172\\
		33.9	1.060660172\\
		34	1.060660172\\
		34.1	1.060660172\\
		34.2	1.060660172\\
		34.3	1.060660172\\
		34.4	1.060660172\\
	};
	\addlegendentry{\small Ref.}
	
	\addplot [color=mycolor1]
	table[row sep=crcr]{%
		0.1	1.83982029626864\\
		0.2	1.80601453043901\\
		0.3	1.32596188766457\\
		0.4	0.947478915075501\\
		0.5	0.601295480209761\\
		0.6	0.305297464911052\\
		0.7	0.212155081034186\\
		0.8	0.407706894278465\\
		0.9	0.637508174379923\\
		1	0.847592678673748\\
		1.1	1.0300342825008\\
		1.2	1.18345874771209\\
		1.3	1.30885411246205\\
		1.4	1.40862031419317\\
		1.5	1.48596622160557\\
		1.6	1.54443433645169\\
		1.7	1.58701679572182\\
		1.8	1.61563802819304\\
		1.9	1.63111069465776\\
		2	1.63344342905436\\
		2.1	1.62230867172\\
		2.2	1.59752592560777\\
		2.3	1.55948086948279\\
		2.4	1.5095581449933\\
		2.5	1.45118590615588\\
		2.6	1.38915295283191\\
		2.7	1.32974309661261\\
		2.8	1.27887108684303\\
		2.9	1.23959927700327\\
		3	1.21116102877903\\
		3.1	1.18850083587941\\
		3.2	1.16340147207588\\
		3.3	1.12681434365204\\
		3.4	1.07098392297863\\
		3.5	0.986716571053736\\
		3.6	0.867118382481238\\
		3.7	0.712254824936288\\
		3.8	0.531155666741813\\
		3.9	0.345071546947453\\
		4	0.212223192228844\\
		4.1	0.250235686544454\\
		4.2	0.388589991294517\\
		4.3	0.522849674014448\\
		4.4	0.620952625859762\\
		4.5	0.67047241659892\\
		4.6	0.671294853687683\\
		4.7	0.635799425128343\\
		4.8	0.58352816390381\\
		4.9	0.536108102218038\\
		5	0.511855683905862\\
		5.1	0.52111947207499\\
		5.2	0.561630006280047\\
		5.3	0.621440927115051\\
		5.4	0.686306988649622\\
		5.5	0.74417155827836\\
		5.6	0.786522700465935\\
		5.7	0.807840253825507\\
		5.8	0.80491532971039\\
		5.9	0.776551846589574\\
		6	0.723485376218447\\
		6.1	0.648710738524635\\
		6.2	0.558649529685915\\
		6.3	0.465758439814731\\
		6.4	0.391946740744701\\
		6.5	0.371858835012497\\
		6.6	0.430061472338487\\
		6.7	0.5507344499391\\
		6.8	0.70345737693319\\
		6.9	0.865178859462256\\
		7	1.01906943054619\\
		7.1	1.15193304216863\\
		7.2	1.25380609878833\\
		7.3	1.31849559855331\\
		7.4	1.34422148453089\\
		7.5	1.33402382589756\\
		7.6	1.29520668726288\\
		7.7	1.23749135134141\\
		7.8	1.1749303729648\\
		7.9	1.12487043197084\\
		8	1.10514231798719\\
		8.1	1.12768091625749\\
		8.2	1.1899159985232\\
		8.3	1.27573506122091\\
		8.4	1.36354510494622\\
		8.5	1.43114545946426\\
		8.6	1.45968485040158\\
		8.7	1.43701660323695\\
		8.8	1.35922254921945\\
		8.9	1.23062176520908\\
		9	1.06255058782718\\
		9.1	0.87106247540982\\
		9.2	0.675436033138164\\
		9.3	0.498795142351474\\
		9.4	0.370825836209204\\
		9.5	0.32257595008032\\
		9.6	0.349393739337227\\
		9.7	0.407816204088164\\
		9.8	0.466391142189233\\
		9.9	0.514625353836784\\
		10	0.552800495791983\\
		10.1	0.584661541937273\\
		10.2	0.612690969217738\\
		10.3	0.635861540471139\\
		10.4	0.649966782661352\\
		10.5	0.64957271772325\\
		10.6	0.630212088688237\\
		10.7	0.590207160536169\\
		10.8	0.532297332894032\\
		10.9	0.465946378681457\\
		11	0.410991136289093\\
		11.1	0.397930770971703\\
		11.2	0.44854817608036\\
		11.3	0.553535745878505\\
		11.4	0.688849500916853\\
		11.5	0.835591183161049\\
		11.6	0.981562507362268\\
		11.7	1.11886545008164\\
		11.8	1.24271089322095\\
		11.9	1.35068470459434\\
		12	1.44206869516537\\
		12.1	1.51713939686177\\
		12.2	1.57633582087296\\
		12.3	1.61983261424192\\
		12.4	1.64745777467907\\
		12.5	1.65885231896399\\
		12.6	1.65377606311253\\
		12.7	1.63283544353833\\
		12.8	1.59771319237995\\
		12.9	1.55133309998422\\
		13	1.49827318792888\\
		13.1	1.44446060050068\\
		13.2	1.39444818803805\\
		13.3	1.3492316297005\\
		13.4	1.30606331268244\\
		13.5	1.25919681655773\\
		13.6	1.20034980715637\\
		13.7	1.12161243450867\\
		13.8	1.01838779311804\\
		13.9	0.891173601605982\\
		14	0.746491621926334\\
		14.1	0.598153365520259\\
		14.2	0.471083675356314\\
		14.3	0.40574247558618\\
		14.4	0.434290275433566\\
		14.5	0.534738809625658\\
		14.6	0.663627969705903\\
		14.7	0.793409183517872\\
		14.8	0.908904111780203\\
		14.9	1.00036560072086\\
		15	1.06157490211979\\
		15.1	1.08983162768423\\
		15.2	1.08602354138974\\
		15.3	1.05466368916835\\
		15.4	1.00461094469722\\
		15.5	0.948095934470079\\
		15.6	0.897485612814106\\
		15.7	0.862179347153598\\
		15.8	0.846892267933326\\
		15.9	0.851728246605849\\
		16	0.873338039335244\\
		16.1	0.906152137978874\\
		16.2	0.943143253040495\\
		16.3	0.976943545247686\\
		16.4	1.00160129200212\\
		16.5	1.0138658282752\\
		16.6	1.01401802228117\\
		16.7	1.0055809983324\\
		16.8	0.99506418439561\\
		16.9	0.990117326319024\\
		17	0.997847993016097\\
		17.1	1.02416117137614\\
		17.2	1.07209831534842\\
		17.3	1.13943876507408\\
		17.4	1.21825794186007\\
		17.5	1.29746719473302\\
		17.6	1.3661330196853\\
		17.7	1.41556291737323\\
		17.8	1.4394600642137\\
		17.9	1.43336244986269\\
		18	1.39457956317706\\
		18.1	1.32199044012575\\
		18.2	1.21742476702516\\
		18.3	1.08636049596158\\
		18.4	0.93801164018007\\
		18.5	0.785909732789144\\
		18.6	0.650470888710698\\
		18.7	0.56227446944132\\
		18.8	0.553117916462847\\
		18.9	0.625260746348854\\
		19	0.748401687325752\\
		19.1	0.891572989172383\\
		19.2	1.0355506574429\\
		19.3	1.16972664927228\\
		19.4	1.28856751333485\\
		19.5	1.39003220780887\\
		19.6	1.47420554221298\\
		19.7	1.54224903785443\\
		19.8	1.59527831048615\\
		19.9	1.63385457794006\\
		20	1.65790670956303\\
		20.1	1.66692166927655\\
		20.2	1.66058835397758\\
		20.3	1.63921680552913\\
		20.4	1.60377101036341\\
		20.5	1.55630140224303\\
		20.6	1.49968929925104\\
		20.7	1.43711033177721\\
		20.8	1.37151101318889\\
		20.9	1.30537179524905\\
		21	1.24076140280327\\
		21.1	1.17893522944605\\
		21.2	1.11931915820819\\
		21.3	1.05891807687463\\
		21.4	0.993223890685945\\
		21.5	0.915161664660438\\
		21.6	0.816710479598891\\
		21.7	0.693404044625034\\
		21.8	0.548613053896091\\
		21.9	0.391330703533763\\
		22	0.235537737327208\\
		22.1	0.1206599923061\\
		22.2	0.163902360272061\\
		22.3	0.289922736563028\\
		22.4	0.423092231020605\\
		22.5	0.555005622867863\\
		22.6	0.684286248624782\\
		22.7	0.807487173643376\\
		22.8	0.917978149823399\\
		22.9	1.00794138222802\\
		23	1.07086689708548\\
		23.1	1.10313287871853\\
		23.2	1.10477622838331\\
		23.3	1.0797462184102\\
		23.4	1.03571147813054\\
		23.5	0.983295368529133\\
		23.6	0.934763452681443\\
		23.7	0.902384344460764\\
		23.8	0.895559588554494\\
		23.9	0.917461304478521\\
		24	0.963186037338489\\
		24.1	1.02005782424718\\
		24.2	1.07186573353665\\
		24.3	1.10533550997916\\
		24.4	1.11404622392753\\
		24.5	1.09722629637211\\
		24.6	1.05746306044379\\
		24.7	0.998749411625432\\
		24.8	0.925086772042789\\
		24.9	0.841082196685669\\
		25	0.750977148343626\\
		25.1	0.658451133379291\\
		25.2	0.567886113269773\\
		25.3	0.486098946449311\\
		25.4	0.422493875847566\\
		25.5	0.386685908731096\\
		25.6	0.382428129675989\\
		25.7	0.402925830889384\\
		25.8	0.43398370168148\\
		25.9	0.461812826623717\\
		26	0.479056968302083\\
		26.1	0.486613428771485\\
		26.2	0.491982318484282\\
		26.3	0.504854287168821\\
		26.4	0.530766138487715\\
		26.5	0.566011253809601\\
		26.6	0.599493441616287\\
		26.7	0.618481308446797\\
		26.8	0.612757903424905\\
		26.9	0.57606872168626\\
		27	0.506113892918312\\
		27.1	0.403951161958245\\
		27.2	0.274622828427949\\
		27.3	0.132546897022462\\
		27.4	0.107175667658738\\
		27.5	0.269064721749459\\
		27.6	0.455627108007354\\
		27.7	0.645154855796114\\
		27.8	0.829897280880241\\
		27.9	1.00351898889571\\
		28	1.16011380596141\\
		28.1	1.29404039771574\\
		28.2	1.40008205611219\\
		28.3	1.47460065176562\\
		28.4	1.51680842361682\\
		28.5	1.53005502640432\\
		28.6	1.5212658499565\\
		28.7	1.49962260010348\\
		28.8	1.47332856561458\\
		28.9	1.44641362929982\\
		29	1.41754128235867\\
		29.1	1.38270250277753\\
		29.2	1.33840615231431\\
		29.3	1.28502788375311\\
		29.4	1.227196537193\\
		29.5	1.17018942005839\\
		29.6	1.11708658361213\\
		29.7	1.06809988663641\\
		29.8	1.02154457926134\\
		29.9	0.974519110189201\\
		30	0.92251307838614\\
		30.1	0.859173257690873\\
		30.2	0.776985826185194\\
		30.3	0.672948359438142\\
		30.4	0.550941246892769\\
		30.5	0.421735114828927\\
		30.6	0.306897501580642\\
		30.7	0.249004709890264\\
		30.8	0.283506596015456\\
		30.9	0.373194661322906\\
		31	0.471526880350938\\
		31.1	0.556577771243798\\
		31.2	0.618874920659921\\
		31.3	0.655888721582647\\
		31.4	0.67059461733111\\
		31.5	0.670432584545111\\
		31.6	0.66591066485313\\
		31.7	0.66864310539761\\
		31.8	0.687710248288867\\
		31.9	0.725979063920186\\
		32	0.780259180811012\\
		32.1	0.84462486377913\\
		32.2	0.913320324671018\\
		32.3	0.981762626124092\\
		32.4	1.04635111561469\\
		32.5	1.10408789845905\\
		32.6	1.15242683623165\\
		32.7	1.18940615142439\\
		32.8	1.21390272099493\\
		32.9	1.22584053458925\\
		33	1.23605876577283\\
		33.1	1.2534650564049\\
		33.2	1.29002565931772\\
		33.3	1.3363707669244\\
		33.4	1.36917776760484\\
		33.5	1.36162079301752\\
		33.6	1.29181312658547\\
		33.7	1.14916019166375\\
		33.8	0.936827387578089\\
		33.9	0.669253933866548\\
		34	0.367704088089317\\
		34.1	0.0694413697673586\\
		34.2	0.265641953198754\\
		34.3	0.547659271945259\\
		34.4	0.792155121415273\\
	};
	\addlegendentry{\small Actual}
	
	\addplot [thick, color=mycolor3, dashed]
	table[row sep=crcr]{%
		0	1.5\\
		35	1.5\\
	};
	\addlegendentry{\small Limits}
	
	\addplot [thick, color=mycolor3, dashed, forget plot]
	table[row sep=crcr]{%
		0	0\\
		35	0\\
	};
\end{axis}

\end{tikzpicture}%

%% file: tex/conclusion.tex
\section{Conclusion and Outlook}\label{Sec:Conclusion}
In this work, we define the Kinematic Orienteering Problem and benchmark exact and heuristically obtained solutions against optimal solutions of the DOP. For flight time estimation, we present an improved analytical approach to calculate time-optimal multidimensional trajectories with bounded acceleration and velocity and demonstrate why the state-of-the-art approach is not valid in general. Further, we show that the obtained overall trajectories can precisely be tracked by a modern MPC-based UAV flight controller. To our knowledge, this constitutes the first approach that enables time-optimal mission planning for multirotor UAVs with consideration of their full physical capabilities. In future work, we will improve the computational performance and quality of our heuristic solution approach and investigate the KOP in three-dimensional scenarios.

%% file: bib/bibliography.bib
@article{Meyer.2021,
	author={Meyer, F. and Glock, K}, 
	year={2021}, 
	title= {Trajectory-based {T}raveling {S}alesman {P}roblem for multirotor {UAV}s}, 
	journal={International Conference on Distributed Computing in Sensor Systems}
}

@article{Fountoulakis.2020,
	author={Fountoulakis, E. and Paschos, G. S. and Pappas, N. },
	year={2020},
	title={{UAV} trajectory optimization for time constrained applications},
	journal={IEEE Networking Letters},
	pages={136-139}, 
	volume={2},
	number={3}
}

@article{Golden1987, 
	author={Golden, B. and Levy, L. and Vohra, R.}, 
	year={1987}, 
	title={The Orienteering Problem}, 
	journal={Naval Research Logistics},
	volume={34}, 
	pages={"307-318}
}

@article{Luukkonen.2011, 
	author={Luukkonen, T.},
	year={2011},
	title={Modelling and control of quadcopter},
	journal={Independent research project in applied mathematics, Espoo: Alto University}
}

@article{Tzorakoleftherakis.2018,
	author={Tzorakoleftherakis, E and Murphey, T. D.},
	year={2018},
	title={Iterative sequential action control for stable, model-based control of nonlinear systems},
	journal={IEEE Transactions on Automatic Control}
}

@article{Miller.1960,
	author = {Miller, C. E. and Tucker, A. W. and Zemlin, R. A.},
	year = {1960},
	title = {Integer programming formulation of {T}raveling {S}alesman {P}roblems},
	journal = {Journal of the ACM},
	volume = {7},
	pages = {326 - 329}
}

@article{Otto.2018,
	author={Otto, A. and Agatz, N. and Campbell, J. and Golden, B. and Pesch, E.},
	year={2018},
	title={Optimization approaches for civil applications of unmanned aerial vehicles ({UAV}s) or aerial drones: {A} survey},
	journal={Networks},
	volume={72},
	pages={411-458}
}

@article{Gao.2018, 
	author={Gao, F. and Wu, W. and Pan, J. and Zhou, B. and Shen, S.},
	year={2018},
	title={Optimal Time Allocation for Quadrotor Trajectory Generation}, 
	journal={International Conference on Intelligent Robots and Systems (IROS)}
}

@article{Henchey.2020,
	author={Henchey, M. and Rosen, S.},
	year={2020},
	title={Emerging approaches to support dynamic mission planning: survey and recommendations for future research},
	journal={Journal of Defense Modelling and Simulation: Applications, Methodology, Technology}	
}

@article{Dubins.1957,
	author={Dubins, L. E.}, 
	year={1957},
	title={On curves of minimal length with a constraint on average curvature, and with prescribed initial and terminal positions and tangents},
	journal={American Journal of Mathematics}, 
	volume={79},
	number={3},
	pages={497-516}
}

@article{Penicka.2017,
	author={Penicka, R. and Faigl, J. and Vana, P. and Saska. M.},
	year={2017},
	title={Dubins orienteering problem}, 
	journal={IEEE Robotics and Automation Letters},
	volume={2}, 
	pages={1210-1217}
}

@article{Faigl.2019,
	author = {Faigl, J. and Vana, P. and Penicka, R.},
	year = {2019},
	title={Multi-vehicle close enough orienteering problem with {B}ézier curves for multi-rotor aerial vehicles},
	journal={International Conference on Robotics and Automation}
}

@article{Faigl.2018,
	author = {Faigl, J. and Vana, P.},
	year = {2019},
	title={Surveillance planning with {B}ézier curves},
	journal={IEEE Robotics and Automation Letters},
	volume = {3},
	number={2},
	pages={750-757}
}

@article{Mueller.2013,
	author={Mueller, M. W. and  D'Andrea, R.},
	year={2013},
	title={A model predictive controller for quadrotor state interception}, 
	journal={European Control Conference}	
}

@article{Sundar.2022,
		author = {Sundar, K. and Sanjeevi, S. and Montez, C.},
		title = {{A} branch-and-price algorithm for a team orienteering problem with fixed-wing drones},
		journal = {EURO Journal on Transportation and Logistics},
		volume = {11},
		pages = {100070},
		year = {2022}
}

@article{Richter.2016,
	author={Charles, R. and Bry, A. and Roy, N.},
	year={2016},
	title={Polynomial trajectory planning for aggressive quadrotor flight in dense indoor environments},
	journal={Robotic Research},
	pages={649-666}
}

@inbook{Kamel.2017,
	author={Kamel, M. and Stastny, T. and Alexis, K. and Siegward, R.},
	title={{M}odel predictive control for trajectory tracking of unmanned aerial vehicles using robot operating system},
	bookTitle={Robot Operating System (ROS): The Complete Reference  (Volume 2)},
	year={2017}, 
	publisher={Springer International Publishing},
	pages={3-39},
	editor={Koubaa, A.},
}

@article{Mellinger.2011,
	author={Mellinger, D. and Kumar, V.}, 
	year={2011}, 
	title={Minimum snap trajectory generation and control for quadrotors},
	journal={IEEE International Conference on Robotics and Automation}
}

@article{Beul.2016, 
	author={Beul, M. and Behnke, S.},
	year={2016},
	title={Analytical time-optimal trajectory generation and control for multirotors},
	journal={International Conference on Unmanned Aircraft Systems (ICUAS)}
}

@article{Beul.2017, 
	author={Beul, M. and Behnke, S.},
	year={2017},
	title={Fast full state trajectory generation for multirotors},
	journal={International Conference on Unmanned Aircraft Systems (ICUAS)}
}

@article{Tsiligirides.1984,
	author={Tsiligirides, T.}, 
	year={1984},
	title={Heuristic methods applied to orienteering},
	journal={J. Oper. Res. Soc.}, 
	volume={35},
	number={9}, 
	pages={797-809}
}

@article{Hehn.2015,
	author={Hehn, M. and D'Andrea, R.}, 
	year={2015},
	title={Real-Time Trajectory Generation for Quadrotors},
	journal={IEEE Transactions on Robotics}, 
	volume={31},
	issue={4}, 
	pages={877-892}
}

@article{Mueller2.2013,
	author={Mueller, M. W. and Hehn, M. and D'Andrea, R.}, 
	year={2013},
	title={{A} computationally efficient algorithm for state-to-state quadrotor trajectory generation and feasibility verification},
	journal={International Conference on Intelligent Robots and Systems (IROS)}
}

@book{Pontryagin.1987, 
	author={Pontryagin, L.}, 
	year={1987}, 
	title={Mathematical Theory of Optimal Processes}, 
	publisher={Taylor \& Francis}
}

@inbook{Pisinger.2018,
	author={Pisinger, D. and Ropke, S.},
	title={{L}arge {N}eighborhood {S}earch},
	bookTitle={Handbook of Metaheuristics},
	year={2018}, 
	publisher={Springer International Publishing},
	pages={99-127},
	editor={Gendreau, M. and Potvin, J.-Y.},
}
